\documentclass[10pt,twocolumn,letterpaper]{article}

\usepackage{cvpr}
\cvprfinalcopy % *** Uncomment this line for the final submission
\usepackage{times}
\usepackage[utf8]{inputenc} % allow utf-8 input
\usepackage[T1]{fontenc}    % use 8-bit T1 fonts
\usepackage{url}            % simple URL typesetting
\usepackage{booktabs}       % professional-quality tables
\usepackage{amsfonts}       % blackboard math symbols
\usepackage{nicefrac}       % compact symbols for 1/2, etc.
\usepackage{microtype}      % microtypography
\usepackage{mathtools}
\usepackage{booktabs}
\usepackage{multirow}
\usepackage{color}
\usepackage{pifont}
\usepackage{wrapfig}
\usepackage[pagebackref=true,breaklinks=true,letterpaper=true,colorlinks,bookmarks=false]{hyperref}
\usepackage{adjustbox}
\usepackage{xcolor}
\usepackage{soul}

\newcommand{\cmark}{\ding{51}}
\newcommand{\xmark}{\ding{55}}

\def\cvprPaperID{1970}

\definecolor{brown}{RGB}{128, 76, 37}
\definecolor{yellow}{RGB}{252, 196, 3}

\ifcvprfinal\fi
\begin{document}
\title{
Hyperparameter-Free 
Out-of-Distribution Detection\\ 
Using Softmax of Scaled Cosine Similarity
}

\author{
Engkarat Techapanurak$^{1}$~~~~~ Masanori Suganuma$^{2,1}$ ~~~~~Takayuki Okatani$^{1,2}$\\
$^1$Graduate School of Information Sciences, Tohoku University ~~~~ $^2$RIKEN Center for AIP\\
Sendai, 980-8579, Japan \\
{\tt\small \{engkarat, suganuma, okatani\}@vision.is.tohoku.ac.jp}
}

\maketitle

\begin{abstract} 
The ability to detect out-of-distribution (OOD) samples is vital to secure the reliability of deep neural networks in real-world applications. Considering the nature of OOD samples, detection methods should not have hyperparameters that need to be tuned depending on incoming OOD samples. However, most of the recently proposed methods do not meet this requirement, leading to compromised performance in real-world applications. In this paper, we propose a simple, hyperparameter-free method based on softmax of scaled cosine similarity. It resembles the approach employed by modern metric learning methods, but it differs in details; the differences are essential to achieve high detection performance. We show through experiments that our method outperforms the existing methods on the evaluation test recently proposed by Shafaei et al., which takes the above issue of hyperparameter dependency into account. We also show that it achieves at least comparable performance to other methods on the conventional test, where their hyperparameters are chosen using explicit OOD samples. Furthermore, it is computationally more efficient than most of the previous methods, since it needs only a single forward pass.
\end{abstract}

\section{Introduction}
\label{sec:intro}

It is widely recognized that deep neural networks tend to show unpredictable behaviors for {\em out-of-distribution} (OOD) samples, i.e., samples coming from a different distribution from that of the training samples. They often give high confidence (i.e., high softmax value) to OOD samples, not only to 
{\em in-distribution} (ID) samples (i.e., test samples from the same distribution as the training samples). Therefore, it has been a major research topic to detect OOD samples in classification performed by deep neural networks; many methods have been proposed so far \cite{hendrycks2016baseline, liang2017enhancing, lee2018simple, vyas2018out, shalev2018out, malinin2018predictive, Yu2019UnsupervisedOD}. 

    \begin{figure}[]
        \centering
        \includegraphics[width=0.97\columnwidth]{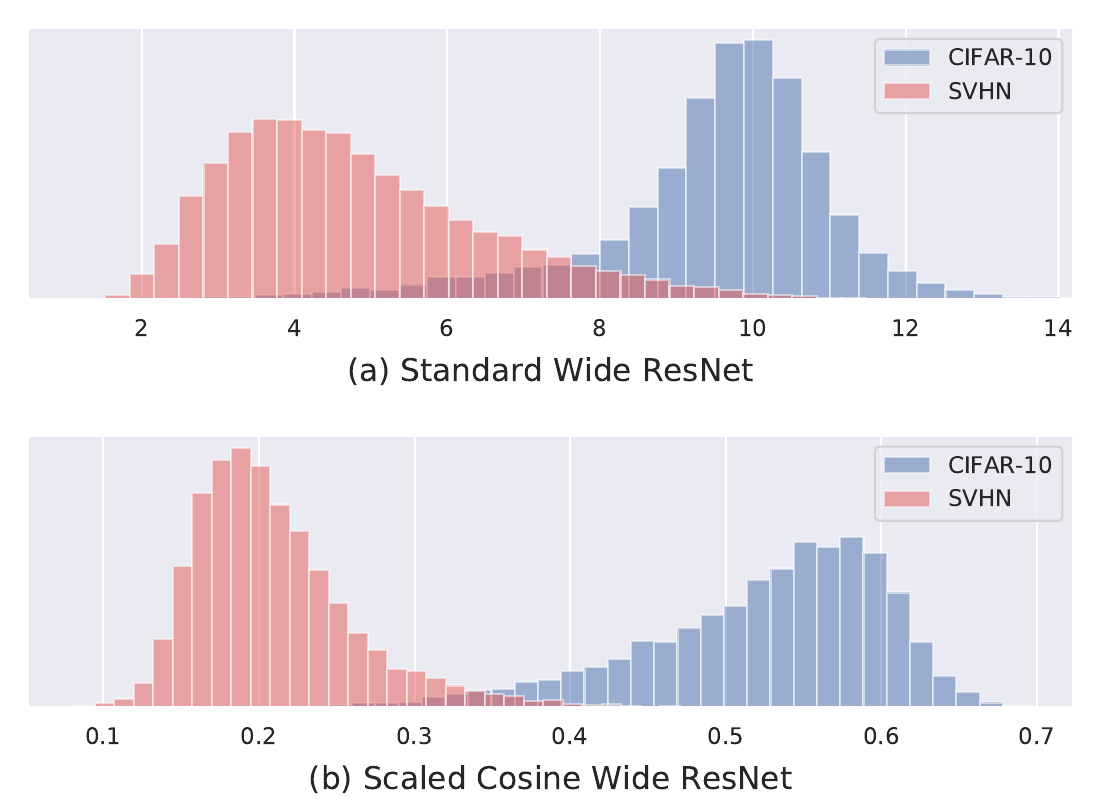}
        \caption{
        Histograms of the maximum of logits over classes. ID (CIFAR-10, 10,000 samples) is in blue color and OOD (SVHN, 26,032) is in red. Histograms are normalized. 
        (a) A standard network (Wide ResNet). (b) Proposed network (the same net with the last layer replaced by the proposed layer of Fig.~\ref{fig::network_diagram}). }
        \label{fig::histogram}
    \end{figure}

A problem with the existing methods, especially those considered to be  state-of-the-art, is that they have hyperparameters specific to OOD detection. They determine these hyperparameters using a certain amount of OOD samples as `validation' data; that is, these studies assume the availability of (at least a small amount of) OOD samples.
This assumption, however, is unlikely to hold true in practice; considering the definition of OOD, it is more natural to assume its distribution to be unknown. Even when the assumption is indeed wrong, it will be fine if OOD detection performance is insensitive to the choice of the hyperparamters, more rigorously, {\em if the hyperparameters tuned on the assumed OOD samples generalize well to incoming OOD samples we encounter in practice.}
However, a recent study \cite{shafaei2018biased} indicates that this is not the case, concluding that none of the existing methods is ready to use, especially for the tasks with  high-dimensional data space, e.g., image classification.

In this paper, we propose a novel method that is free of any hyperparameters associated with OOD detection. 
As there is no need to access OOD samples to determine hyperparameters, the proposed method is free from the above issue. 
We show that it outperforms the existing methods by a large margin on the recently proposed test \cite{shafaei2018biased}, which takes the above issue of hyperparameter dependency into account. It also shows at least comparable performance to the state-of-the-art methods on the conventional test. 

The proposed method is based on the use of softmax of scaled cosine similarity for modeling class probabilities.
It can be used with any networks; only their output layers need to be changed. Training is performed by the standard method, i.e., minimizing a cross-entropy loss on the ID classification task.
As shown in Fig.~\ref{fig::histogram}, the cosine similarities computed at the last layer exhibit more distinguishable responses between ID and OOD samples than their counterpart (i.e., the logits) in the standard networks. Although it comes at the cost of a small loss of ID classification accuracy, this can be rectified by using the proposed network only for OOD detection and a standard network for ID classification.

In addition to having no hyperparameter, the proposed method is superior to the existing methods in terms of computational efficiency. It needs only a single forward propagation, whereas the existing methods perform more complicated computation, e.g., input perturbation which requires backpropagation for each input \cite{liang2017enhancing, lee2018simple, vyas2018out}.

Although the proposed method resembles the approach employed in recent metric learning methods \cite{ranjan2017l2,liu2017sphereface,wang2017normface,wang2018cosface,deng2018arcface},
it has several differences in technical details, in addition to the difference in purpose, from the metric learning methods. We show through experiments that the technical differences matter for OOD detection performance.

\section{Related Work}
\subsection{Uncertainty of Prediction}
  It is known that when applied to classification tasks, deep neural networks often exhibit overconfidence for unseen inputs. 
  Many studies have been conducted to find a solution to this issue. A popular approach is to evaluate uncertainty of a prediction and use it as its reliability measure. There are many studies on this approach, most of which are based on the framework of Bayesian neural networks or its approximation 
  \cite{gal2017dropout, lakshminarayanan2017simple, azizpour2018bayesian, gast2018lightweight}. It is reported that  predicted uncertainty is useful for real-world applications \cite{kendall2017what, leibig2017leveraging, devries2018leveraging}. However, it is still an open problem to accurately evaluate uncertainty.
  There are also studies on calibration of confidence scores \cite{guo2017calibration, kuleshov2018accurate, subramanya2017confidence}.
  Some studies propose to build a meta system overseeing the classifier that can estimate the reliability of its prediction \cite{scheirer2011meta, chen2018confidence}. 

\subsection{Out-of-distribution (OOD) Detection}
  
\subsubsection{Detection Methods}

  A more direct approach to the above issue is OOD detection. A baseline method that thresholds confidence score, i.e., the maximum softmax output, is evaluated in \cite{hendrycks2016baseline}. This study presents a design of experiments for evaluation of OOD detection methods, which has been employed in the subsequent studies. Since then, many studies have been conducted. It should be noted that these methods have hyperparameters for OOD detection, which need to be determined in some way. Some studies assume a portion of OOD samples to be given and regard them as a `validation' set, by which the hyperparemters are determined.
  
  ODIN \cite{liang2017enhancing} applies perturbation with a constant magnitude $\epsilon$ to an input $x$ in the direction of increasing the confidence score (i.e., the maximum softmax) and then uses the increased score in the same way as the baseline. An observation behind this procedure is that such perturbation tends to increase confidence score more for ID samples than for OOD samples. Rigorously, $x$ is perturbed to increase a temperature-scaled softmax value.
  Thus, ODIN has two hyperparameters $\epsilon$ and the softmax temperature. 
  In the experiments reported in \cite{liang2017enhancing}, $\epsilon$ as well as the temperature are determined by using a portion of samples from a target OOD dataset; this is done for each pair of ID and OOD datasets.
  
  The current state-of-the-art of OOD detection is achieved by the methods \cite{vyas2018out,lee2018simple} employing input perturbation similar to ODIN. It should be noted that there are many studies with different motivations, such as generative models \cite{lee2017training, ren2019likelihood}, a prior distribution \cite{malinin2018predictive}, robustification by training networks to predict word embedding of class labels \cite{shalev2018out}, pretraining of networks \cite{hendrycks2018deep, hendrycks2019using}, and batch-wise fine-tuning \cite{Yu2019UnsupervisedOD}.
  
  In \cite{vyas2018out}, a method that employs an ensemble of networks and similar input perturbation is proposed, achieving the state-of-the-art performance. In the training step of this method, ID classes are split into two sets, one of which is virtually treated as ID classes and the other as OOD classes. A network is then trained so that the entropy for the former samples is minimized while that for the latter samples is maximized. Repeating this for different $K$ splits of classes yields $K$ leave-out classifiers (i.e., networks). At test time, an input $x$ is given to these $K$ networks, whose outputs are summed to calculate ID class scores and an OOD score, where $x$ is perturbed with magnitude $\epsilon$ in the direction of minimizing the entropy. In the experiments, 
  $\epsilon$, the temperature, and additional hyperparameters
  are determined by selecting a particular dataset (i.e., iSUN \cite{xu2015turkergaze}) as the OOD dataset, and OOD detection performance on different OOD datasets is evaluated. 
  
  In \cite{lee2018simple}, another method is proposed, which models layer activation over ID samples with class-wise Gaussian distributions. It uses the induced Mahalanobis distances to class centroids for conducting the classification as well as OOD detection. It employs logistic regression integrating information from multiple layers and input perturbation similar to ODIN, which possesses several hyperparameters.
  For their determination, it is suggested to use explicit OOD samples, as in ODIN \cite{liang2017enhancing}. Another method is additionally suggested to avoid this potentially unrealistic assumption, which is to create adversarial examples for ID samples \cite{goodfellow2015explaining} and use them as OOD samples, determining the hyperparameters.
  However, even this method is not free of hyperparameters; the creation of adversarial examples needs at least one (i.e., perturbation magnitude). 
  It is not discussed how to choose it in their paper.

\subsubsection{Evaluation Methods}  
\label{sec:evaltest}

Most of the recent studies employ the following evaluation method \cite{hendrycks2016baseline}. Specifying a pair of ID and OOD datasets (e.g., CIFAR-10 for ID and SVHN for OOD), it measures accuracy of distinguishing the OOD samples and ID samples. 
As the task is detection, appropriate  metrics are used, such as accuracy at true positive rate (TPR) $=95\%$, area under the ROC curve (AUROC), and under the precision-recall curve (AUPR).
As is noted in Sec.~\ref{sec:intro}, most of the existing methods assume the availability of OOD samples and use them to determine their hyperparameters. Note that these OOD samples are selected from the {\em true} OOD dataset specified in this evaluation method. We will refer to this {\em one-vs-one} evaluation.

Recently, Shafaei et al. have raised a concern about the dependency of the existing methods on the explicit knowledge of the true OOD dataset, and proposed a novel evaluation method that aims at measuring the practical performance of OOD detection \cite{shafaei2018biased}.
It assumes an ID dataset and multiple OOD datasets ${\cal D}=\{D_1,\ldots\}$ for evaluation. Then, the evaluation starts with choosing one dataset $D_i\in {\cal D}$ and use the samples from it to determine the hyperparameters of the method under evaluation; it then evaluates its detection accuracy when regarding each of the other datasets in ${\cal D}$ (i.e., ${\cal D}\backslash D_i$)  as the OOD dataset, reporting the average accuracy over ${\cal D}\backslash D_i$. Note that this test returns the accuracy for each dataset in ${\cal D}$ (used for the assumed OOD dataset).
We will refer to this {\em less-biased} evaluation.

\subsection{Cosine Similarity}   
   
  The proposed method employs softmax of scaled cosine similarities instead of ordinary softmax of logits. 
  A similar approach has already been employed in recent studies of metric learning, such as $L_2$-constrained softmax \cite{ranjan2017l2}, SphereFace \cite{liu2017sphereface}, NormFace \cite{wang2017normface}, CosFace \cite{wang2018cosface}, ArcFace \cite{deng2018arcface}, etc. Although it may seem rather straightforward to apply these methods to OOD detection, to the authors' knowledge, there is no study that has tried this before.
 
  These metric learning methods are identical in that they use cosine similarity to improve a few issues with the ordinary softmax. They differ in i) if and how the weight $\mathbf{w}$ or the feature $\mathbf{f}$ of the last layer are normalized and ii) if and how margins are used with the cosine similarity that (further) encourage maximization of inter-class variance and minimization of intra-class variance. According to this categorization, our method is the most similar to NormFace \cite{wang2017normface}, in which both $\mathbf{w}$ and $\mathbf{f}$ are normalized and no margin is utilized. However, our method still differs from NormFace that it uses only a single fully-connected layer to compute the cosine similarity and it predicts scale $s$ multiplied with the inputs to softmax, as will be described below. 

\section{Proposed Method}

\begin{figure}[t]
        \centering
        \includegraphics[width=0.9\linewidth]{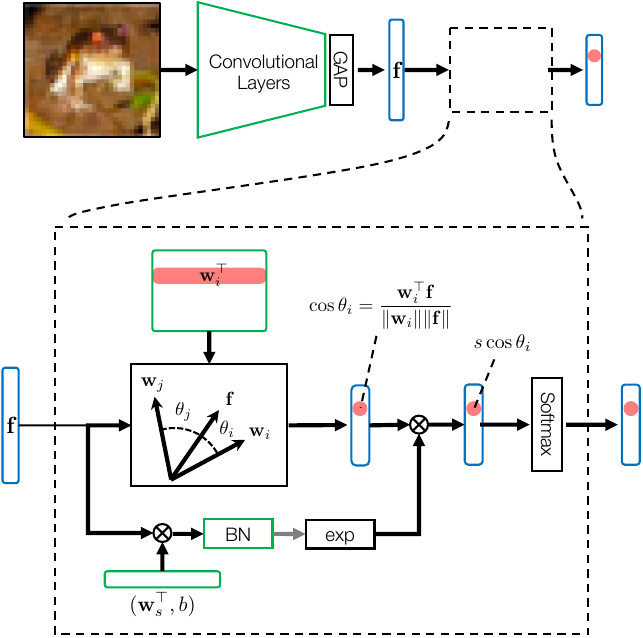}
        \smallskip
        \caption{Proposed architecture for computing scaled cosine similarity. The blank box in the upper row, which is a fully-connected layer(s) in a standard CNN, is replaced with the architecture shown below.}
        \label{fig::network_diagram}
    \end{figure}

    Figure \ref{fig::network_diagram} shows the diagram of the proposed method. Details are described below. 

  \subsection{Softmax of Scaled Cosine Similarity}

    The standard formulation of multi-class classification is to make the network predict class probabilities for an input, and use cross-entropy loss to evaluate the correctness of the prediction. 
    The predicted class probabilities are obtained by applying softmax to the linear transform $\mathbf{W}\mathbf{f}+\mathbf{b}$ of the activation or feature $\mathbf{f}$ of the last layer, and then  the loss is calculated assuming 1-of-$K$ coding of the true class $c$ as
    \begin{equation}
        \mathcal{L} = - \log\frac{e^{\mathbf{w}^\top_c\mathbf{f} + b_c}}{\sum_{i=1}^C e^{\mathbf{w}_i^\top\mathbf{f} + b_i}},
        \label{eq:ce}
    \end{equation}
    where 
    $\mathbf{W}=[\mathbf{w}_1,\ldots,\mathbf{w}_C]^\top$ and $\mathbf{b}=[b_1,\ldots,b_C]^\top$. 
    
    Metric learning attempts to learn feature space suitable for the purpose of open-set classification, e.g., face verification. Unlike earlier methods employing triplet loss \cite{wang2014learning, schroff2015facenet} and contrastive loss \cite{hadsell2006dimensionality, sun2016sparsifying}, recent methods \cite{wang2017normface, wang2018cosface, deng2018arcface}  modify the loss (\ref{eq:ce}) and minimize the cross entropy loss as with the standard multi-class classification. The main idea is to use the cosine of the angle between the weight $\mathbf{w}_i$ and the feature $\mathbf{f}$ as a class score. Specifically, $\cos\theta_i\equiv \mathbf{w}_i^\top\mathbf{f} / (\lVert \mathbf{w}_i \rVert \lVert \mathbf{f}\rVert)$ is used instead of the logit $\mathbf{w}_i^\top\mathbf{f}+b_i$ in (\ref{eq:ce}); then a new loss is given as
    \begin{equation}
         \mathcal{L} = - \log\frac{e^{\cos{\theta}_c}}{\sum_{i=1}^C e^{\cos{\theta}_i}}.
    \end{equation}
    
    The behavior of softmax, i.e., how soft its maximum operation will be, depends on the distribution of its inputs, which can be controlled by a scaling parameter of the inputs, called temperature $T$. This parameter is used for several purposes \cite{hinton2015distilling,guo2017calibration}. In metric learning methods, it is employed to widen the range $[-1,1]$ 
    of $\cos\theta_i$'s inputted to softmax; specifically, all the input cosine $\cos\theta_i$'s are scaled by a parameter $s(=1/T)$, revising the above loss as 
    \begin{equation}
    \label{eq:targetloss}
         \mathcal{L} = - \log\frac{e^{s\cos{\theta}_c}}{\sum_{i=1}^C e^{s\cos{\theta}_i}}.
    \end{equation}

\subsection{Predicting the Scaling Parameter} \label{sec:predscale}

    In most of the metric learning methods employing similar loss functions, the scaling parameter $s$ in (\ref{eq:targetloss}) is treated as a hyperparameter and its value is chosen in a validation step. An exception is NormFace \cite{wang2017normface}, in which $s$ is treated as a learnable parameter like ordinary weights; $s$ is adjusted by computing the gradient with respect to it.
    Besides these two methods, there is yet another way of determining $s$, which is to predict it from $\mathbf{f}$ together with class scores. We empirically found that this performs the best.
    Among several ways of computing $s$ from $\mathbf{f}$, the following works the best: 
        \begin{equation}
        \label{eqn:preds}
            s = \exp{\{\mathrm{BN}(\mathbf{w}_s^\top\mathbf{f}+b_s)\}},
        \end{equation}
    where $\mathrm{BN}$ is batch normalization \cite{ioffe2015batch}, and $\mathbf{w}_s$ and $b_s$ are the weight and bias of the added branch to predict $s$.

  \subsection{Design of the Output Layer}
  
  \label{sec:outputlayer}
  
In the aforementioned studies of metric learning, 
ResNets are employed as a base network and are modified to implement the softmax of cosine similarity. 
Modern CNNs like ResNets
are usually designed to have a single fully-connected (FC) layer between the final pooling layer (i.e., global average pooling) and the network output. 
As ReLU activation function is applied to the inputs of the pooling layer, if we use the last FC layer for computing cosine similarity (i.e., treating its input as $\mathbf{f}$ and its weights as $\mathbf{w}_i$'s), then the elements of $\mathbf{f}$ take only non-negative values. Thus, the metric learning methods add an extra single FC layer on top of the FC layer and use the output of the first FC layer as $\mathbf{f}$, making $\mathbf{f}$ (after normalization) distribute on the whole hypersphere. In short, the metric learning methods employ two FC layers at the final section of the network. 

However, we found that for the purpose of OOD detection, having two fully-connected layers does not perform better than simply using the output of the final pooling layer as $\mathbf{f}$. Details will be given in our experimental results. Note that in the case of a single FC layer, as $\mathbf{f}$ takes only non-negative values, $\mathbf{f}$ resides in the first quadrant of the space, which is very narrow subspace comparative to the entire space.

To train the modified network, we use a standard method.
In our experiments, we employ SGD with weight decay as the optimizer, as in the previous studies of OOD detection \cite{liang2017enhancing, lee2018simple, shalev2018out, vyas2018out}. In several studies of metric learning \cite{wang2018cosface, deng2018arcface, liu2016large}, weight decay is also employed on all the layers of networks. However, it may have different effects on the last layer of the network employing cosine similarity, where weights are normalized and thus its length does not affect the loss. In our experiments, we found that it works better when we do not apply weight decay to the last layer. 

   \subsection{Detecting OOD samples}

    Detecting OOD samples is performed in the following way. Given an input $x$, our network computes $\cos\theta_i$ ($i=1,\ldots,C$).
    Let $i_{\mathrm max}$ be the index of the maximum of these cosine values. We use $\cos\theta_{i_{\mathrm max}}$ for distinguishing ID and OOD samples. To be specific, setting a threshold, we declare $x$ is an OOD sample if $\cos\theta_{i_{\mathrm max}}$ is lower than it. Otherwise, we classify $x$ into the class $i_{\mathrm max}$ with the predicted probability $e^{s \cos{\theta_{i_{\mathrm max}}}}/\sum e^{s \cos{\theta_i}}$.

\section{Experimental Results}

\subsection{Experimental Settings} \label{experimental_settings}
  
We conducted experiments to evaluate the proposed method and compare it with existing methods. 

\subsubsection{Evaluation Methods}

We employ the one-vs-one and less-biased evaluation methods explained in Sec.~\ref{sec:evaltest}. The major difference between the two is in the assumption of prior knowledge about OOD datasets, which affects the determination of the hyperparameters of the OOD detection methods under evaluation. 
Note therefore that {\em the difference does not matter for our method}, as it does not need any hyperparameter; it only affects the other compared methods. 

\paragraph{One-vs-one evaluation} This evaluation assumes one ID and one OOD datasets. A network is trained on the ID dataset and each method attempts to distinguish ID and OOD samples using the network. Each method may use a fixed number of samples from the specified OOD datasets for its hyperparameter determination. We followed the experimental configurations commonly employed in the previous studies \cite{liang2017enhancing, lee2018simple, vyas2018out}. 

\paragraph{Less-biased evaluation}
This evaluation uses one ID and many OOD datasets. Each method may access one of the OOD datasets to determine its hyperparameters but its evaluation is conducted on the task of distinguishing the ID samples and samples from each of the other OOD datasets. We followed the study of Shafaei et al. \cite{shafaei2018biased} with slight modifications. First, we use AUROC instead of detection accuracy for evaluation metrics (additionally, accuracy at TPR$=95\%$ and AUPR-IN in the supplementary material), as we believe that they are better metrics for detection tasks, and they are employed in the one-vs-one evaluation. Second, we add more OOD datasets to those used in their study to further increase the effectiveness and practicality of the evaluation.
    
\subsubsection{Tasks and Datasets} 

We use CIFAR-10/100 for the target classification tasks in all the experiments. Using them as ID datasets, we use the following OOD datasets in one-vs-one evaluation: TinyImageNet (cropped and resized) \cite{deng2009imagenet}, LSUN (cropped and resized) \cite{yu15lsun},
iSUN \cite{xu2015turkergaze},\footnote{Datasets are available at \url{https://github.com/facebookresearch/odin}.} SVHN \cite{netzer2011reading} 
and Food-101 \cite{bossard14food101}
For the less-biased evaluation, we additionally use STL-10 \cite{Coates2011stl10}, MNIST \cite{lecun1998mnist}, NotMNIST, and Fashion MNIST \cite{xiao2017fmnist}. As for STL-10 and Food-101, we resize their images to $32\times32$ pixels.

\paragraph{Remark} {\color{black} We found that the cropped images of TinyImageNet and LSUN that are provided by the GitHub repository of \cite{liang2017enhancing}, which are employed in many recent 
studies, have a black frame of two-pixel width around them; see the supplementary material for details. Although we are not sure if this is intentional, considering that the frame will make OOD detection easier, we use two versions with/without the black frame in our experiments; the frame-free version is indicated by `$*$' in what follows. In the main paper, we show mainly results on the frame-free versions. Those on the original versions are shown in the supplementary material, although it does not affect our conclusion.
}
    
\subsubsection{Networks and Their Training} 

For networks, we employ the two CNNs commonly used in the previous studies, i.e., \textit{Wide ResNet} and \textit{DenseNet} as the base networks. Following \cite{liang2017enhancing}, we use WRN-28-10 and DenseNet-BC having 100 layers with growth rate 12. The former is trained with batch size $=128$ for 200 epochs with weight decay $=0.0005$, and the latter is trained with batch size $=64$ for 300 epochs with weight decay $=0.0001$. Dropout is not used in the both networks. We employ a learning rate schedule, where the learning rate starts with 0.1 and decreases by $1/10$ at 50\% and 75\% of the training steps.
  
The proposed method modifies the final layer and the loss of the base networks. Table \ref{table:classification_results} shows comparisons between the base networks and their modified version. The numbers are an average over five runs and their standard deviations are shown in parenthesis. It is seen that the modification tends to lower classification accuracy by a small amount. If this difference does matter, one may use the proposed network only for OOD detection and the standard network for ID classification.

    \begin{table}[]
    \caption{Performance of the base networks and their modified versions for the proposed method for the task of classification of ID (in-distribution) samples.}
    \label{table:classification_results}
    \begin{center}
    \begin{small}
    \begin{sc}
    \resizebox{0.7\columnwidth}{!}{
    \begin{tabular}{cccc}
    \toprule
    \multirow{2}{*}{Network} & \multirow{2}{*}{In-Dist} & \multicolumn{2}{c}{Testing Accuracy} \\ 
    \cmidrule(l){3-4} 
    & & Standard & Cosine \\
    \midrule
    \multirow{2}{*}{DenseBC}   
        & CIFAR-10 &  95.11(0.10) & 94.92(0.04) \\
        & CIFAR-100 & 76.97(0.24) & 75.65(0.12) \\ 
    \midrule
    \multirow{2}{*}{WRN-28-10} 
        & CIFAR-10 & 95.99(0.09) & 95.72(0.05) \\
        & CIFAR-100 & 81.04(0.37) & 78.53(0.28) \\ 
    \bottomrule
    \end{tabular}
    }
    \end{sc}
    \end{small}
    \end{center}
    \end{table}

\begin{figure*}[t]
\centering
\includegraphics[width=1.0\textwidth]{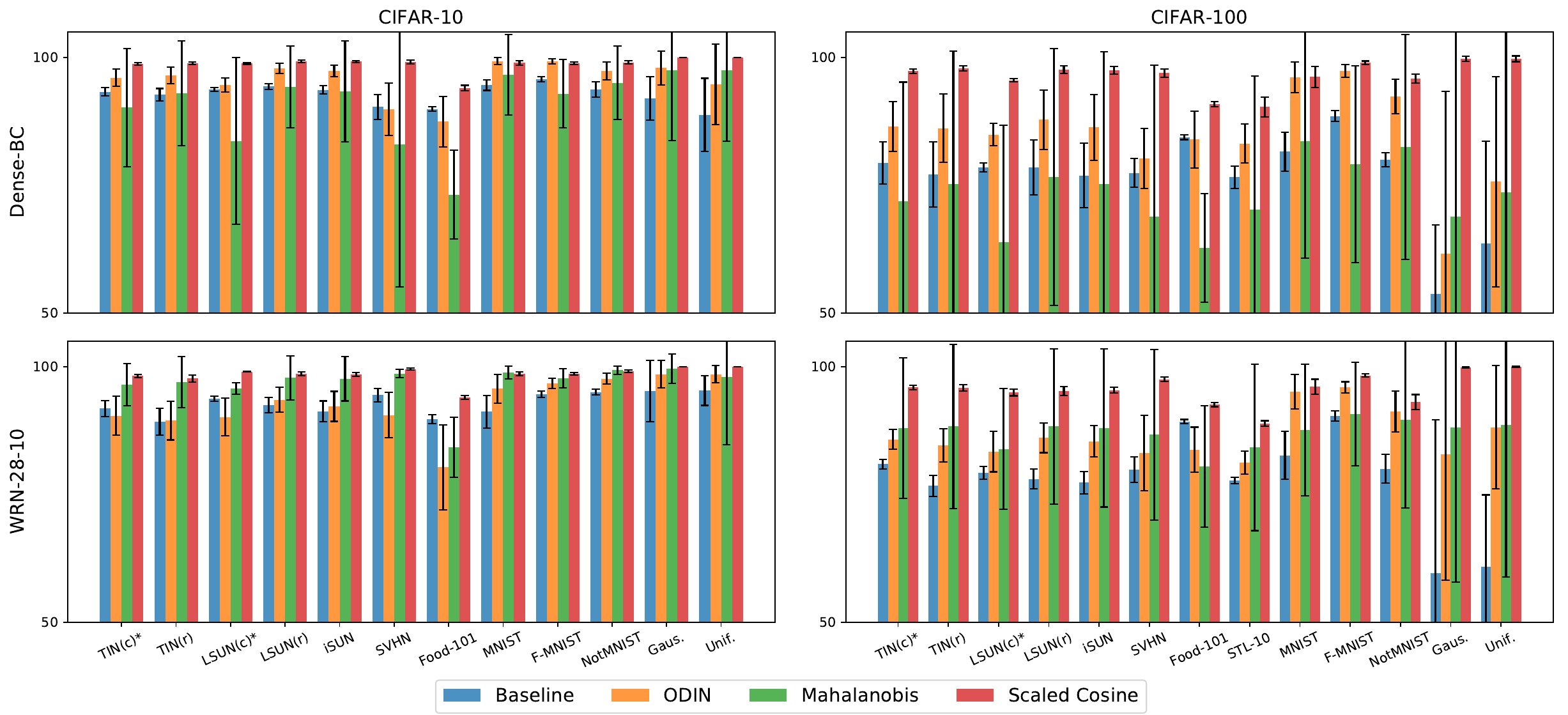}
\caption{OOD detection performance (AUROC) measured by the less-biased evaluation \cite{shafaei2018biased} for the baseline method  \cite{hendrycks2016baseline}, ODIN, \cite{liang2017enhancing} and the Mahalanobis detector \cite{lee2018simple}, and the proposed one (denoted as `Scaled Cosine'). Other metrics, i.e., accuracy at TPR=$95\%$ and AUPR-IN,  are reported in the supplementary material.}
\label{fig::AUROC_less_biased}
\end{figure*}
 
\subsubsection{Compared Methods}
    
    The methods we compare are as follows: the baseline method \cite{hendrycks2016baseline}, ODIN \cite{liang2017enhancing}, Mahalanobis detector \cite{lee2018simple}\footnote{We used the publicly available code: \url{https://github.com/pokaxpoka/deep_Mahalanobis_detector}}, and leave-out ensemble \cite{vyas2018out}. 
    The last two methods are reported to achieve the highest performance in the case of a single network and multiple networks, respectively. We conduct experiments separately with the first three and the last one due to the difference in settings. We report comparisons with the first three in the main paper and those with the leave-out ensemble in the supplementary material for lack of space.

    All these methods (but the baseline) have hyperparameters for OOD detection. For ODIN and the Mahalanobis detector, we follow the authors' methods \cite{liang2017enhancing,lee2018simple} to determine them using a portion of the true OOD dataset. 
    For the leave-out ensemble (comparisons in the supplementary material), we use the values of detection accuracy from its paper \cite{vyas2018out}, in which the authors use a specific OOD dataset (iSUN) for hyperparameter determination.

\subsection{Comparison by Less-biased Evaluation}
    
We first show the performance of the four methods, i.e., the baseline, ODIN, the Mahalanobis detector, and ours, measured by the less-biased evaluation method. Figure \ref{fig::AUROC_less_biased} shows their performance in AUROC\footnote{A complete table including other metrics, i.e., accuracy at TPR$=95\%$ and AUPR-IN, is shown in the supplementary material.}. The dataset names shown along the horizontal axis indicate the assumed OOD datasets, using which the hyperparameters are determined in ODIN and the Mahalanobis detector; all the other datasets are used to evaluate the OOD detection performance. As the detection performance can be computed for any pairs of ID dataset (CIFAR-10 or CIFAR-100) and one of the other OOD datasets, each bar indicates their mean and the associated error bar indicate their standard deviation.
    
It is seen from Fig.~\ref{fig::AUROC_less_biased} that the proposed method consistently achieves better performance than others. It is noted that the Mahalanobis detector, which shows the state-of-the-art performance in  the conventional (i.e., one-vs-one) evaluation, shows unstable behaviors; the mean of AUROC tends to vary significantly and the standard deviation is very large depending on the dataset used for hyperparameter determination. The same observation applies  to ODIN. 

\begin{figure}[]
    \centering
  \includegraphics[width=8cm]{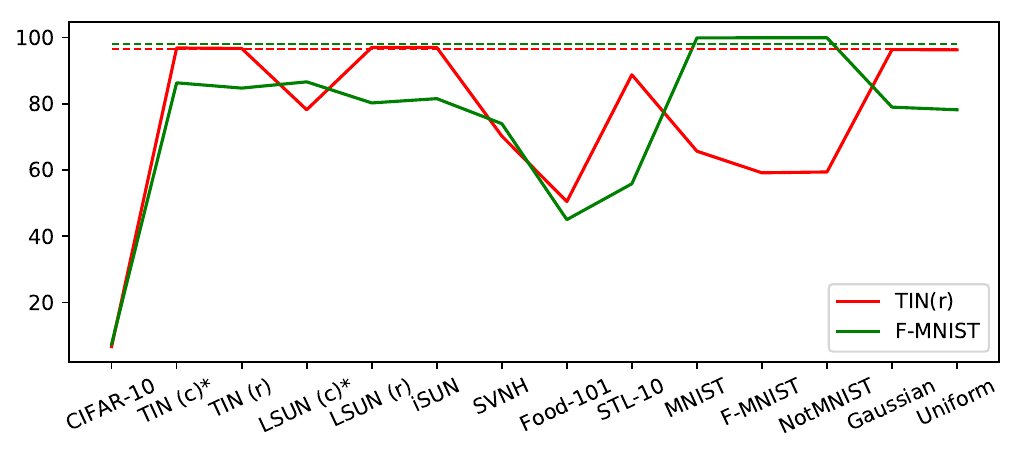}
  \caption{Dependency of detection performance (AUROC) on the assumed OOD datasets (whose names are given in the horizontal axis) used for determining hyperparameters. Mahalanobis detector (solid lines) \cite{lee2018simple} and our method (broken lines). CIFAR-100 is used as ID and either TIN(r) (in red color) or F-MNIST (in green color) is used as true OOD. WRN-28-10 is used for the network. Our method does not have hyperparameters and thus is independent of the assumed OOD dataset.}
  \label{fig::Tuning_Maha}
\end{figure}

This clearly demonstrates the issue with these methods, that is, their performance is dependent on the choice of the hyperparameters. On the other hand, the proposed method performs consistently for all the cases. This is also confirmed from Fig.~\ref{fig::Tuning_Maha}, which shows a different plot of the same experimental result; it shows AUROC for a {\em single} OOD dataset instead of the {\em mean} over multiple OOD datasets shown in Fig.~\ref{fig::AUROC_less_biased}. It is seen that the performance of the Mahalanobis detector varies a lot depending on the assumed OOD dataset. Additionally, it can be seen that the dataset yielding the highest performance differs for different true OOD datasets; iSUN, LSUN(r), or Gaussian etc. is the best for detecting TIN(r) as OOD, whereas MNIST or notMNIST is the best for detecting F-MNIST as OOD.

\subsection{Comparison by One-vs-one Evaluation}

We then show the comparison of the same four methods in the one-vs-one evaluation, which is employed in the majority of the previous studies. 
We ran each method five times from the training step, where the network weights are initialized randomly, and report the mean and standard deviation here. 
Table \ref{table:single_ood_result2} shows the results. It is observed that the proposed method achieves better or at least competitive performance to the others. 
When using DenseNet-BC, the proposed method 
consistently achieves higher performance than the Mahalanobis detector on almost all the datasets. 
    
    \begin{table}[]
    \caption{OOD detection performance of the four methods measured by conventional one-vs-one evaluation.}
    \vskip -0.2in
    \label{table:single_ood_result2}
    \begin{center}
    \begin{small}
    \begin{sc}
    \resizebox{1.0\columnwidth}{!}{
    \begin{tabular}{ccccccc}
    \toprule
    & \multirow{2}{*}{ID} & \multirow{2}{*}{OOD} & \multicolumn{4}{c}{AUROC}  \\ 
    \cmidrule{4-7}
    & & & Base \cite{hendrycks2016baseline} & ODIN \cite{liang2017enhancing} & Maha \cite{lee2018simple} & Cosine \\
    \midrule
    \multirow{18}{*}{\rotatebox[origin=c]{90}{Dense-BC}} & 
    \multirow{9}{*}{\rotatebox[origin=c]{90}{CIFAR-10}} & 
        TIN (c) & 94.90(0.43) & 98.79(0.32) & 94.48(1.19) & \textbf{98.89(0.24)} \\
    & & TIN (c)* & 93.26(0.85) & 96.67(0.97) & 97.36(0.39) & \textbf{98.74(0.23)} \\
    & & TIN (r) & 92.67(1.23) & 97.20(1.17) & \textbf{98.91(0.23)} & 98.82(0.29) \\
    & & LSUN (c) & 95.57(0.20) & 98.48(0.14) & 89.06(3.21) & \textbf{99.09(0.12)} \\
    & & LSUN (c)* & 93.72(0.39) & 96.41(0.52) & 93.63(0.69) & \textbf{98.83(0.18)} \\
    & & LSUN (r) & 94.28(0.52) & 98.43(0.49) & 99.00(0.23) & \textbf{99.19(0.22)} \\
    & & iSUN & 93.62(0.83) & 97.92(0.71) & 98.95(0.21) & \textbf{99.20(0.19)} \\
    & & SVNH & 90.28(2.47) & 95.11(0.48) & 98.89(0.37) & \textbf{99.11(0.36)} \\
    & & Food-101 & 89.87(0.44) & 92.06(0.71) & 80.38(3.83) & \textbf{93.98(0.54)} \\
    \cmidrule(l){2-7}
    & \multirow{9}{*}{\rotatebox[origin=c]{90}{CIFAR-100}}& 
        TIN (c) & 83.70(4.00) & 94.48(3.21) & 92.97(1.63) & \textbf{97.90(0.29)} \\
    & & TIN (c)* & 79.32(4.14) & 88.54(4.27) & 93.18(0.39) & \textbf{97.31(0.45)} \\
    & & TIN (r) & 77.07(6.35) & 88.14(6.92) & 96.81(0.27) & \textbf{97.82(0.53)} \\
    & & LSUN (c) & 82.92(0.59) & 94.72(0.59) & 91.65(2.96) & \textbf{96.73(0.31)} \\
    & & LSUN (c)* & 78.46(0.91) & 87.89(1.13) & 85.44(1.85) & \textbf{95.52(0.32)} \\
    & & LSUN (r) & 78.44(5.41) & 90.38(4.76) & 97.00(0.15) & \textbf{97.59(0.75)} \\
    & & iSUN & 76.89(6.28) & 88.27(6.49) & 97.04(0.10) & \textbf{97.45(0.73)} \\
    & & SVNH & 77.36(2.83) & 91.60(0.73) & 96.48(0.68) & \textbf{96.90(0.79)} \\
    & & Food-101 & 84.38(0.48) & \textbf{90.82(0.60)} & 67.14(1.39) & 90.79(0.49) \\
    \midrule
    \multirow{18}{*}{\rotatebox[origin=c]{90}{WRN-28-10}}&
    \multirow{9}{*}{\rotatebox[origin=c]{90}{CIFAR-10}}& 
        TIN (c) & 93.86(0.90) & 95.88(1.01) & 95.99(1.04) & \textbf{98.35(0.32)} \\
    & & TIN (c)* & 91.79(1.57) & 92.17(2.19) & \textbf{98.50(0.11)} & 98.17(0.33) \\
    & & TIN (r) & 89.21(2.65) & 90.60(3.21) & \textbf{99.15(0.18)} & 97.65(0.66) \\
    & & LSUN (c) & 95.41(0.26) & 97.20(0.15) & 92.65(1.33) & \textbf{99.19(0.07)} \\
    & & LSUN (c)* & 93.67(0.50) & 95.08(0.42) & 96.90(0.35) & \textbf{98.98(0.07)} \\
    & & LSUN (r) & 92.45(1.48) & 94.48(1.70) & \textbf{99.37(0.13)} & 98.59(0.34) \\
    & & iSUN & 91.22(2.05) & 93.25(2.43) & \textbf{99.29(0.10)} & 98.48(0.36) \\
    & & SVNH & 94.43(1.30) & 93.34(3.60) & 99.28(0.09) & \textbf{99.52(0.24)} \\
    & & Food-101 & 89.71(0.90) & 89.18(2.37) & 90.43(1.54) & \textbf{93.95(0.41)} \\
    \cmidrule(l){2-7}
    & \multirow{9}{*}{\rotatebox[origin=c]{90}{CIFAR-100}} & 
        TIN (c) & 84.47(1.24) & 91.72(1.10) & 92.58(2.60) & \textbf{96.76(0.34)} \\
    & & TIN (c)* & 80.90(0.90) & 87.08(1.29) & \textbf{96.45(0.30)} & 95.91(0.42) \\
    & & TIN (r) & 76.67(2.03) & 86.28(2.43) & \textbf{97.82(0.13)} & 95.84(0.67) \\
    & & LSUN (c) & 81.91(1.31) & 91.75(0.44) & 80.48(1.14) & \textbf{96.09(0.62)} \\
    & & LSUN (c)* & 79.17(1.25) & 88.06(0.46) & 91.13(0.52) & \textbf{94.92(0.65)} \\
    & & LSUN (r) & 78.00(1.95) & 87.90(1.83) & \textbf{97.80(0.15)} & 95.18(0.86) \\
    & & iSUN & 77.29(2.15) & 87.07(2.00) & \textbf{97.66(0.14)} & 95.39(0.55) \\
    & & SVNH & 79.82(2.49) & 93.46(1.05) & \textbf{97.96(0.49)} & 97.52(0.41) \\
    & & Food-101 & 89.25(0.40) & 90.76(0.35) & 91.15(0.66) & \textbf{92.53(0.38)} \\
    \bottomrule
    \end{tabular}}
    \end{sc}
    \end{small}
    \end{center}
    \vskip -0.2in
    \end{table}

  \subsection{Ablation Study} \label{ablation_study}

    Although the proposed method employs softmax of cosine similarity equivalent to metric learning methods, there are differences in detailed designs, even compared with the most similar NormFace \cite{wang2017normface}. To be specific, they are   
    the scale prediction (referred to as {\em Scale} in Table \ref{table:ablation_study}), the use of a single FC layer instead of two FC layers ({\em Single FC}), and non-application of weight decay to the last FC layer ({\em w/o WD}). To see their impacts on performance, we conducted an ablation study, in which WRN-28-10 is used for the base network and TinyImageNet (r) is chosen for an OOD dataset.
         
    Table \ref{table:ablation_study} shows the results. Row 1 shows the results of the baseline method \cite{hendrycks2016baseline}, which are obtained in our experiments. 
    Row 2  shows the results obtained by incorporating the scale prediction in the standard networks; to be specific, $s$ predicted from $\mathbf{f}$ according to (\ref{eqn:preds}) is multipled with logits as $s\cdot (\mathbf{w}_i\mathbf{f}+b_i)$ $(i=1,\ldots,C)$, which are then normalized by softmax to yield the cross-entropy loss. As is shown in Row 2, this simple modification to the baseline boosts the performance, which is surprising.

\begin{figure}[]
        \centering
      \includegraphics[width=5cm]{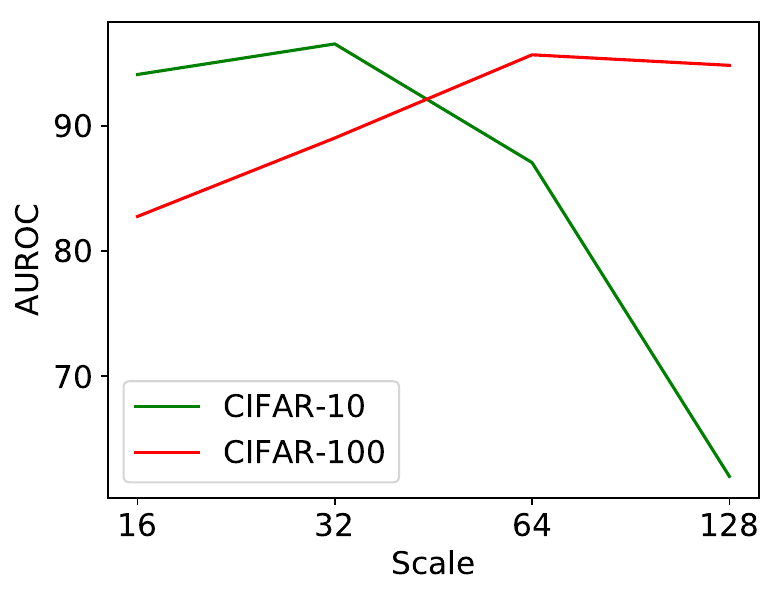}
      \caption{\small Out-of-distribution detection performance (AUROC) when the scale $s$ of cosine similarity is manually specified. }
      \label{fig::scale_auroc}
    \end{figure}

    Row 3 and below show results when cosine similarity is used for OOD detection. Rows 3 to 6 show the results obtained when a fixed value is chosen for $s$. It is observed from this that the application of scaling affects a lot detection performance, and it tends to be sensitive to their choice. This is more clearly seen in Fig.~\ref{fig::scale_auroc} which shows the plot of AUROC values versus scales. This means that, if $s$ is treated as a fixed parameter, it will become a hyperparameter that needs to be tuned for each dataset. Row 7 shows the result when the scale is predicted from $\mathbf{f}$ as in Row 1 but with cosine similarity. It is seen that this provides results comparable to the best case of manually chosen scales. 
    \begin{table}[]
    \caption{Ablation tests for evaluating the contribution of different components (i.e., `Cosine', `Single FC', `Scale', and `w/o WD'; see details from the main text) of the proposed method. AUROCs for detection of OOD samples (TinyImageNet (resized)) are shown.}
    \label{table:ablation_study}
    \begin{center}
    \begin{small}
    \begin{sc}
    \resizebox{1.0\columnwidth}{!}{
    \begin{tabular}{ccccccc} 
    \toprule
    & Cosine & Single FC & Scale & w/o WD & C-10 & C-100 \\
    \midrule
    (1) & \multicolumn{4}{c}{Baseline \cite{hendrycks2016baseline}} & 89.22 & 76.59 \\
    \midrule
    (2) & \xmark & \cmark & Pred & \xmark & 95.74 & 88.70 \\
    (3) & \cmark & \cmark & 16 & \xmark & 94.09 & 82.76 \\
    (4) & \cmark & \cmark & 32 & \xmark & 96.53 & 89.02 \\
    (5) & \cmark & \cmark & 64 & \xmark & 87.06 & 95.66 \\
    (6) & \cmark & \cmark & 128 & \xmark & 62.02 & 94.82 \\
    (7) & \cmark & \cmark & Pred & \xmark & 95.16 & 91.30 \\
    (8) & \textbf{\cmark} & \textbf{\cmark} & \textbf{Pred} & \textbf{\cmark} & \textbf{97.66} & \textbf{95.84} \\
    (9) & \cmark & \xmark & Pred & \xmark & 94.71 & 87.55 \\
    (10) & \cmark & \xmark & Pred & \cmark & 89.90 & 86.96 \\ \bottomrule
    \end{tabular}
    }
    \end{sc}
    \end{small}
    \end{center}
    \end{table}
    
\begin{figure*}[]
\centering
\includegraphics[width=0.905\textwidth]{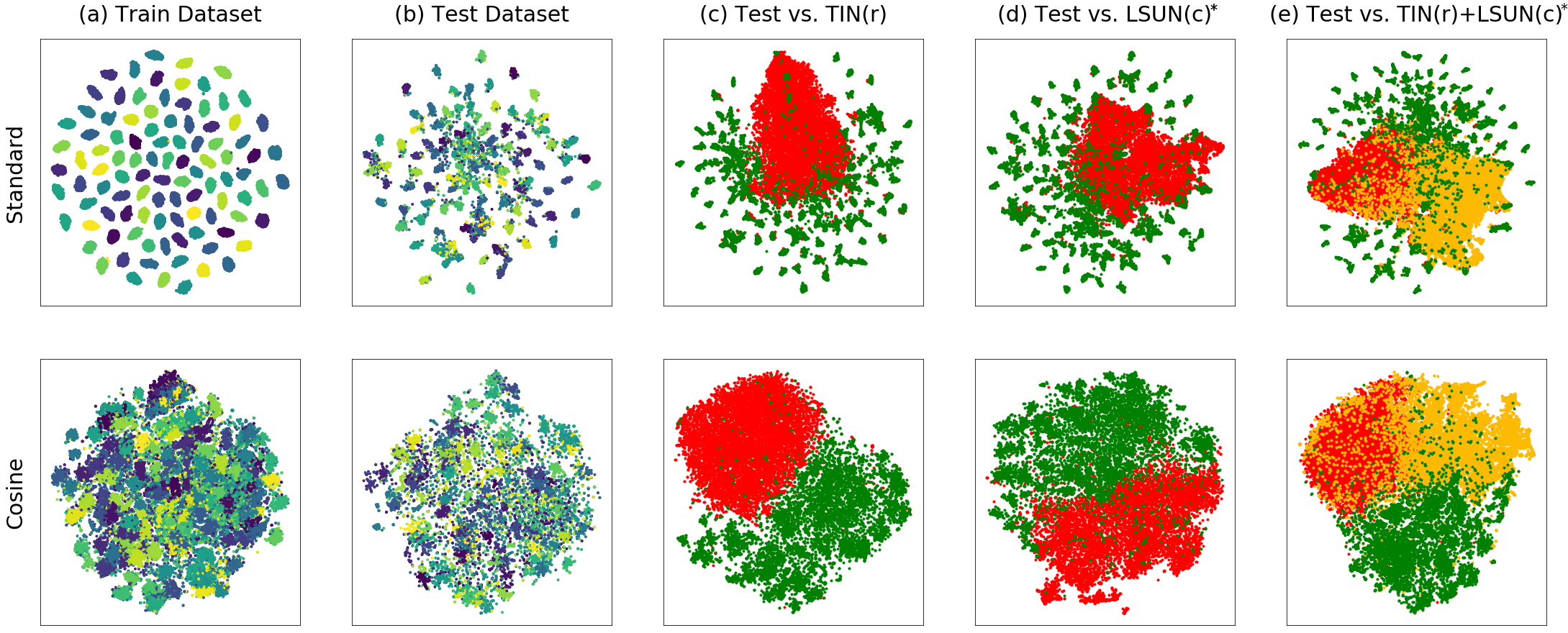}
\caption{The t-SNE plots of various ID/OOD samples in the space of the penultimate layer feature of the two networks trained on CIFAR-100. The upper and lower rows show the standard network and the proposed cosine-similarity network, respectively. From left to right columns: (a) ID (i.e., CIFAR-100) training samples. (b) ID test samples. (c) ID test vs. TinyImageNet(r). (d) ID test vs. LSUN(c)* (e) ID test vs. TinyImageNet(r)+LSUN(c)*. For (a) and (b), different colors indicate the 100 classes. For (c), (d), and (e), green color indicate ID samples and red (and yellow in (e)) color indicate OOD samples. Best viewed in color.
}
\label{fig::tsne_plots}
\end{figure*}
    
    Row 8 shows the results obtained by further stopping application of weight decay to the last layer, which is the proposed method. It is seen that this achieves the best performance for both CIFAR-10 and CIFAR-100. Rows 9 and 10 show the results obtained by the network having two FC layers in its final part, as in the recent metric learning methods. Following the studies of metric learning, we use 512 units in the intermediate layer.
    In this architecture, it is better to employ weight decay in the last layer 
    as with the metric learning methods 
    (i.e., Rows 9 vs 10). In conclusion, these results confirm that the use of cosine similarity as well as all the three components are indispensable to achieve the best performance.

\subsection{Computational Cost}

While the proposed method needs only the standard forward propagation to perform OOD detection, the previous methods, particularly those showing good performance in the one-vs-one evaluation, employ a lot more complicated computation, such as input perturbation \cite{liang2017enhancing, lee2018simple}. 
We measure the computational time that ODIN, the Mahalanobis detector, and ours need to get the results. 
Table \ref{table:calculation_time} shows the average time per batch containing 128 samples.

    \begin{table}[h]
    \caption{Comparison of computational time (per batch of 128 samples) of the three methods. }
    \label{table:calculation_time}
    \begin{center}
    \begin{small}
    \begin{sc}
    \resizebox{5.5cm}{!}{
    \begin{tabular}{cccc}
    \toprule
    \multirow{2}{*}{Network} & \multicolumn{3}{c}{Time (second)} \\
    \cmidrule{2-4} 
     & Maha & ODIN & Cosine  \\
    \midrule
    Dense-BC & 0.67 & 0.19 & \textbf{0.08} \\
    WRN-28-10 & 1.61 & 0.61 & \textbf{0.22} \\
    \bottomrule
    \end{tabular}}
    \end{sc}
    \end{small}
    \end{center}
    \end{table}

\section{Conclusion and Discussions}

In this paper, we have presented a novel method for OOD detection, and experimentally confirmed its superiority to existing approaches. We started our discussion with the observation that existing methods have hyperparameters specific to OOD detection, and their performance can be sensitive to their determination.
The proposed method does not have such hyperparameters. It is based on the softmax of scaled cosine similarity and can be used with any networks by replacing their output layer. Training is performed by the standard method, i.e., minimizing a cross-entropy loss on the target classification task. Although a similar approach has already been employed in metric learning methods, the proposed method has several technical differences, which are essential to achieve high OOD detection performance, as was demonstrated in our ablation test. 

We have shown experimental comparisons between the proposed method and the existing methods using two different evaluation methods, i.e., the less-biased evaluation recently proposed in \cite{shafaei2018biased} and the conventional one-vs-one evaluation. In the former evaluation, which takes the above issue with hyperparameter determination into account, the proposed method shows clear superiority to others. Our method also shows at least comparable performance to them in the conventional evaluation. 
It is also computationally efficient.
These results support the practicality of the proposed method in real-world applications.

An open question is why this method works so well. Although we are conducting a study to answer this question, we have not reached a complete conclusion yet. Instead, as additional evidence of the effectiveness of the proposed approach, we show here the t-SNE plots of the feature spaces learned by a standard network and the proposed network in Fig.~\ref{fig::tsne_plots}. It is observed that in the feature space of the standard network, while samples from each ID class distribute compactly, which may contribute to the accuracy of ID classification, the distribution of OOD samples tends to have substantial overlap with the ID sample distribution. On the other hand, in the feature space of the proposed network, samples of each ID class have comparatively less compact distribution, whereas the OOD sample distribution has smaller overlap with the ID sample distribution, which arguably demonstrates well the superiority of the proposed method.

\bibliographystyle{ieee_fullname}
\bibliography{main}

\begin{thebibliography}{10}\itemsep=-1pt

\bibitem{azizpour2018bayesian}
Hossein Azizpour, Mattias Teye, and Kevin Smith.
\newblock Bayesian uncertainty estimation for batch normalized deep networks.
\newblock In {\em Proceedings of the International Conference on Machine
  Learning}, 2018.

\bibitem{bossard14food101}
Lukas Bossard, Matthieu Guillaumin, and Luc Van~Gool.
\newblock Food-101 -- mining discriminative components with random forests.
\newblock In {\em Proceedings of The European Conference on Computer Vision},
  2014.

\bibitem{chen2018confidence}
Tongfei Chen, Ji{\v{r}}{\'\i} Navr{\'a}til, Vijay Iyengar, and Karthikeyan
  Shanmugam.
\newblock Confidence scoring using whitebox meta-models with linear classifier
  probes.
\newblock In {\em Proceedings of the International Conference on Artificial
  Intelligence and Statistics}, 2018.

\bibitem{Coates2011stl10}
Adam Coates, Honglak Lee, and Andrew~Y. Ng.
\newblock An analysis of single layer networks in unsupervised feature
  learning.
\newblock In {\em Proceedings of the International Conference on Artificial
  Intelligence and Statistics}, 2011.

\bibitem{deng2009imagenet}
Jia Deng, Wei Dong, Richard Socher, Li-Jia Li, Kai Li, and Li Fei-Fei.
\newblock Imagenet: A large-scale hierarchical image database.
\newblock In {\em Proceedings of the IEEE Conference on Computer Vision and
  Pattern Recognition}, 2009.

\bibitem{deng2018arcface}
Jiankang Deng, Jia Guo, Niannan Xue, and Stefanos Zafeiriou.
\newblock Arcface: Additive angular margin loss for deep face recognition.
\newblock In {\em Proceedings of the IEEE Conference on Computer Vision and
  Pattern Recognition}, 2019.

\bibitem{devries2018leveraging}
Terrance DeVries and Graham~W Taylor.
\newblock Leveraging uncertainty estimates for predicting segmentation quality.
\newblock {\em arXiv:1807.00502}, 2018.

\bibitem{gal2017dropout}
Yarin Gal and Zoubin Ghahramani.
\newblock Dropout as a bayesian approximation: Representing model uncertainty
  in deep learning.
\newblock In {\em Proceedings of the International Conference on Machine
  Learning}, 2016.

\bibitem{gast2018lightweight}
Jochen Gast and Stefan Roth.
\newblock Lightweight probabilistic deep networks.
\newblock In {\em Proceedings of the IEEE Conference on Computer Vision and
  Pattern Recognition}, 2018.

\bibitem{goodfellow2015explaining}
Ian~J Goodfellow, Jonathon Shlens, and Christian Szegedy.
\newblock Explaining and harnessing adversarial examples.
\newblock In {\em Proceedings of the International Conference on Learning
  Representations}, 2015.

\bibitem{guo2017calibration}
Chuan Guo, Geoff Pleiss, Yu Sun, and Kilian~Q Weinberger.
\newblock On calibration of modern neural networks.
\newblock In {\em Proceedings of the International Conference on Machine
  Learning}, 2017.

\bibitem{hadsell2006dimensionality}
Raia Hadsell, Sumit Chopra, and Yann LeCun.
\newblock Dimensionality reduction by learning an invariant mapping.
\newblock In {\em Proceedings of the IEEE Conference on Computer Vision and
  Pattern Recognition}, 2006.

\bibitem{hendrycks2016baseline}
Dan Hendrycks and Kevin Gimpel.
\newblock A baseline for detecting misclassified and out-of-distribution
  examples in neural networks.
\newblock In {\em Proceedings of the International Conference on Learning
  Representations}, 2017.

\bibitem{hendrycks2019using}
Dan Hendrycks, Kimin Lee, and Mantas Mazeika.
\newblock Using pre-training can improve model robustness and uncertainty.
\newblock In {\em Proceedings of the International Conference on Machine
  Learning}, 2019.

\bibitem{hendrycks2018deep}
Dan Hendrycks, Mantas Mazeika, and Thomas~G Dietterich.
\newblock Deep anomaly detection with outlier exposure.
\newblock In {\em Proceedings of the International Conference on Learning
  Representations}, 2019.

\bibitem{hinton2015distilling}
Geoffrey Hinton, Oriol Vinyals, and Jeff Dean.
\newblock Distilling the knowledge in a neural network.
\newblock In {\em Advances in Neural Information Processing Systems}, 2014.

\bibitem{ioffe2015batch}
Sergey Ioffe and Christian Szegedy.
\newblock Batch normalization: Accelerating deep network training by reducing
  internal covariate shift.
\newblock In {\em Proceedings of the International Conference on Machine
  Learning}, 2015.

\bibitem{kendall2017what}
Alex Kendall and Yarin Gal.
\newblock What uncertainties do we need in bayesian deep learning for computer
  vision?
\newblock In {\em Advances in Neural Information Processing Systems}, 2017.

\bibitem{kuleshov2018accurate}
Volodymyr Kuleshov, Nathan Fenner, and Stefano Ermon.
\newblock Accurate uncertainties for deep learning using calibrated regression.
\newblock In {\em Proceedings of the International Conference on Machine
  Learning}, 2018.

\bibitem{lakshminarayanan2017simple}
Balaji Lakshminarayanan, Alexander Pritzel, and Charles Blundell.
\newblock Simple and scalable predictive uncertainty estimation using deep
  ensembles.
\newblock In {\em Advances in Neural Information Processing Systems}, 2017.

\bibitem{lecun1998mnist}
Yann LeCun, L{\'e}on Bottou, Yoshua Bengio, Patrick Haffner, et~al.
\newblock Gradient-based learning applied to document recognition.
\newblock In {\em Proceedings of the IEEE}, 1998.

\bibitem{lee2017training}
Kimin Lee, Honglak Lee, Kibok Lee, and Jinwoo Shin.
\newblock Training confidence-calibrated classifiers for detecting
  out-of-distribution samples.
\newblock In {\em Proceedings of the International Conference on Learning
  Representations}, 2018.

\bibitem{lee2018simple}
Kimin Lee, Kibok Lee, Honglak Lee, and Jinwoo Shin.
\newblock A simple unified framework for detecting out-of-distribution samples
  and adversarial attacks.
\newblock In {\em Advances in Neural Information Processing Systems}, 2018.

\bibitem{leibig2017leveraging}
Christian Leibig, Vaneeda Allken, Murat~Se{\c{c}}kin Ayhan, Philipp Berens, and
  Siegfried Wahl.
\newblock Leveraging uncertainty information from deep neural networks for
  disease detection.
\newblock {\em Scientific reports}, 7(1):17816, 2017.

\bibitem{liang2017enhancing}
Shiyu Liang, Yixuan Li, and R Srikant.
\newblock Enhancing the reliability of out-of-distribution image detection in
  neural networks.
\newblock In {\em Proceedings of the International Conference on Learning
  Representations}, 2017.

\bibitem{liu2017sphereface}
Weiyang Liu, Yandong Wen, Zhiding Yu, Ming Li, Bhiksha Raj, and Le Song.
\newblock Sphereface: Deep hypersphere embedding for face recognition.
\newblock In {\em Proceedings of the IEEE Conference on Computer Vision and
  Pattern Recognition}, 2017.

\bibitem{liu2016large}
Weiyang Liu, Yandong Wen, Zhiding Yu, and Meng Yang.
\newblock Large-margin softmax loss for convolutional neural networks.
\newblock In {\em Proceedings of the International Conference on Machine
  Learning}, 2016.

\bibitem{malinin2018predictive}
Andrey Malinin and Mark Gales.
\newblock Predictive uncertainty estimation via prior networks.
\newblock In {\em Advances in Neural Information Processing Systems}, 2018.

\bibitem{netzer2011reading}
Yuval Netzer, Tao Wang, Adam Coates, Alessandro Bissacco, Bo Wu, and Andrew~Y
  Ng.
\newblock Reading digits in natural images with unsupervised feature learning.
\newblock In {\em Advances in Neural Information Processing Systems Workshop on
  Deep Learning and Unsupervised Feature Learning}, 2011.

\bibitem{ranjan2017l2}
Rajeev Ranjan, Carlos~D Castillo, and Rama Chellappa.
\newblock L2-constrained softmax loss for discriminative face verification.
\newblock {\em arXiv:1703.09507}, 2017.

\bibitem{ren2019likelihood}
Jie Ren, Peter~J. Liu, Emily Fertig, Jasper Snoek, Ryan Poplin, Mark~A.
  DePristo, Joshua~V. Dillon, and Balaji Lakshminarayanan.
\newblock Likelihood ratios for out-of-distribution detection.
\newblock In {\em Advances in Neural Information Processing Systems}, 2019.

\bibitem{scheirer2011meta}
Walter~J Scheirer, Anderson Rocha, Ross~J Micheals, and Terrance~E Boult.
\newblock Meta-recognition: The theory and practice of recognition score
  analysis.
\newblock {\em IEEE Transactions on Pattern Analysis and Machine Intelligence},
  33(8):1689--1695, 2011.

\bibitem{schroff2015facenet}
Florian Schroff, Dmitry Kalenichenko, and James Philbin.
\newblock Facenet: A unified embedding for face recognition and clustering.
\newblock In {\em Proceedings of the IEEE Conference on Computer Vision and
  Pattern Recognition}, 2015.

\bibitem{shafaei2018biased}
Alireza Shafaei, Mark Schmidt, and James~J. Little.
\newblock A less biased evaluation of out-of-distribution sample detectors.
\newblock In {\em Proceedings of the British Machine Vision Conference}, 2019.

\bibitem{shalev2018out}
Gabi Shalev, Yossi Adi, and Joseph Keshet.
\newblock Out-of-distribution detection using multiple semantic label
  representations.
\newblock In {\em Advances in Neural Information Processing Systems}, 2018.

\bibitem{subramanya2017confidence}
Akshayvarun Subramanya, Suraj Srinivas, and R~Venkatesh Babu.
\newblock Confidence estimation in deep neural networks via density modelling.
\newblock In {\em Proceedings of IEEE International Conference on Multimedia
  and Expo}, 2017.

\bibitem{sun2016sparsifying}
Yi Sun, Xiaogang Wang, and Xiaoou Tang.
\newblock Sparsifying neural network connections for face recognition.
\newblock In {\em Proceedings of the IEEE Conference on Computer Vision and
  Pattern Recognition}, 2016.

\bibitem{vyas2018out}
Apoorv Vyas, Nataraj Jammalamadaka, Xia Zhu, Dipankar Das, Bharat Kaul, and
  Theodore~L Willke.
\newblock Out-of-distribution detection using an ensemble of self supervised
  leave-out classifiers.
\newblock In {\em Proceedings of the European Conference on Computer Vision},
  2018.

\bibitem{wang2017normface}
Feng Wang, Xiang Xiang, Jian Cheng, and Alan~Loddon Yuille.
\newblock Normface: L2 hypersphere embedding for face verification.
\newblock In {\em Proceedings of the ACM International Conference on
  Multimedia}, 2017.

\bibitem{wang2018cosface}
Hao Wang, Yitong Wang, Zheng Zhou, Xing Ji, Dihong Gong, Jingchao Zhou, Zhifeng
  Li, and Wei Liu.
\newblock Cosface: Large margin cosine loss for deep face recognition.
\newblock In {\em Proceedings of the IEEE Conference on Computer Vision and
  Pattern Recognition}, 2018.

\bibitem{wang2014learning}
Jiang Wang, Yang Song, Thomas Leung, Chuck Rosenberg, Jingbin Wang, James
  Philbin, Bo Chen, and Ying Wu.
\newblock Learning fine-grained image similarity with deep ranking.
\newblock In {\em Proceedings of the IEEE Conference on Computer Vision and
  Pattern Recognition}, 2014.

\bibitem{xiao2017fmnist}
Han Xiao, Kashif Rasul, and Roland Vollgraf.
\newblock Fashion-mnist: a novel image dataset for benchmarking machine
  learning algorithms.
\newblock {\em arXiv:1708.07747}, 2017.

\bibitem{xu2015turkergaze}
Pingmei Xu, Krista~A Ehinger, Yinda Zhang, Adam Finkelstein, Sanjeev~R
  Kulkarni, and Jianxiong Xiao.
\newblock Turkergaze: Crowdsourcing saliency with webcam based eye tracking.
\newblock {\em arXiv:1504.06755}, 2015.

\bibitem{yu15lsun}
Fisher Yu, Yinda Zhang, Shuran Song, Ari Seff, and Jianxiong Xiao.
\newblock Lsun: Construction of a large-scale image dataset using deep learning
  with humans in the loop.
\newblock {\em arXiv:1506.03365}, 2015.

\bibitem{Yu2019UnsupervisedOD}
Qing Yu and Kiyoharu Aizawa.
\newblock Unsupervised out-of-distribution detection by maximum classifier
  discrepancy.
\newblock In {\em Proceedings of the International Conference on Computer
  Vision}, 2019.

\end{thebibliography}

\onecolumn
\appendix

\begin{center}
    \Large\textbf{Supplementary Material to ``Hyperparameter-Free Out-of-Distribution Detection Using Softmax of Scaled Cosine Similarity''}\\
\end{center}

\section{Detailed Results of Less-Biased Evaluation}

    In the main paper, we report results of the less-biased evaluation in Fig.~\ref{fig::AUROC_less_biased}. We show here more detailed results of the same experiments including two additional metrics, i.e., accuracy at TPR$=95\%$ and AUPR-In, in Table \ref{table:OOD_detection_result_less_biased}. In the experiments, we train the network for each case five times with random initial weights. Each line for each metric shows the mean and standard deviation over these five models and also all the datasets except the one specified in the first column, which is the assumed validation dataset used for hyperparameter determination. 

    \begin{table}[h]
    \caption{More detailed results of OOD detection performance measured by the less-biased evaluation. The same four methods as those considered in the main paper are compared. }
    \label{table:OOD_detection_result_less_biased}
    \begin{center}
    \begin{small}
    \begin{sc}
    \resizebox{1.0\textwidth}{!}{
    \begin{tabular}{cccccc}
    \toprule
    Network & ID & OOD & Accuracy at TPR$=95\%$ & AUROC & AUPR-In  \\ \midrule
    & & & \multicolumn{3}{c}{Baseline\cite{hendrycks2016baseline}  /  ODIN\cite{liang2017enhancing}  /  Mahalanobis\cite{lee2018simple}  /  Ours} \\ \midrule
    \multirow{24}{*}{\rotatebox[origin=c]{90}{DenseBC}} & \multirow{12}{*}{\rotatebox[origin=c]{90}{CIFAR-10}} & 
        TIN (c)* & 74.53(1.93) / 87.45(3.49) / 81.94(11.54) / \textbf{94.34(0.61)} & 93.26(0.85) / 96.02(1.64) / 90.16(11.53) / \textbf{98.74(0.23)} & 94.61(0.80) / 95.87(2.20) / 89.54(13.12) / \textbf{98.86(0.19)} \\
    & & TIN (r) & 73.51(2.48) / 88.42(4.02) / 86.52(12.24) / \textbf{94.67(0.60)} & 92.67(1.23) / 96.43(1.64) / 92.97(10.27) / \textbf{98.82(0.29)} & 94.06(1.17) / 96.42(1.92) / 92.36(11.57) / \textbf{98.89(0.26)} \\
    & & LSUN (c)* & 75.35(1.90) / 85.40(2.07) / 74.04(9.96) / \textbf{94.54(0.39)} & 93.72(0.39) / 94.57(1.40) / 83.64(16.27) / \textbf{98.83(0.18)} & 95.06(0.27) / 93.63(2.32) / 83.67(16.41) / \textbf{98.88(0.17)} \\
    & & LSUN (r) & 76.74(1.40) / 92.05(2.55) / 87.07(12.11) / \textbf{95.72(0.56)} & 94.28(0.52) / 97.86(0.97) / 94.21(7.96) / \textbf{99.19(0.22)} & 95.60(0.44) / 97.99(1.01) / 94.22(8.49) / \textbf{99.26(0.20)} \\
    & & iSUN & 75.12(2.08) / 90.60(2.99) / 86.53(12.74) / \textbf{95.62(0.51)} & 93.62(0.83) / 97.33(1.15) / 93.33(9.83) / \textbf{99.20(0.19)} & 95.48(0.66) / 97.69(1.14) / 93.70(9.85) / \textbf{99.33(0.16)} \\
    & & SVNH & 68.84(2.90) / 76.81(7.84) / 81.63(17.75) / \textbf{95.36(0.87)} & 90.28(2.47) / 89.81(5.11) / 82.92(27.86) / \textbf{99.11(0.36)} & 82.90(7.23) / 76.03(11.04) / 78.05(30.89) / \textbf{97.99(0.73)} \\
    & & Food-101 & 67.93(0.59) / 71.83(5.12) / 56.14(4.56) / \textbf{79.79(1.03)} & 89.87(0.44) / 87.43(4.92) / 73.10(8.68) / \textbf{93.98(0.54)} & 83.58(0.96) / 73.45(11.31) / 58.06(11.76) / \textbf{89.99(0.90)} \\
    & & MNIST & 76.35(3.52) / \textbf{95.81(2.11)} / 91.93(11.67) / 95.15(1.51) & 94.54(1.02) / \textbf{99.24(0.68)} / 96.60(7.90) / 98.90(0.49) & 95.98(0.83) / \textbf{99.36(0.55)} / 97.66(5.91) / 99.06(0.42) \\
    & & F-MNIST & 81.65(2.11) / \textbf{95.70(1.25)} / 79.22(9.62) / 94.83(0.71) & 95.76(0.50) / \textbf{99.20(0.47)} / 92.86(6.69) / 98.84(0.21) & 96.76(0.35) / \textbf{99.28(0.39)} / 94.15(6.49) / 98.94(0.17) \\
    & & NotMNIST & 75.95(3.71) / 90.83(5.11) / 89.22(7.87) / \textbf{95.75(1.09)} & 93.72(1.49) / 97.33(1.75) / 94.99(7.17) / \textbf{99.01(0.30)} & 91.90(2.37) / 95.26(2.89) / 89.81(11.58) / \textbf{98.62(0.45)} \\
    & & Gaussian & 63.72(14.12) / 90.27(16.11) / 95.02(10.42) / \textbf{97.50(0.00)} & 91.96(4.29) / 97.92(3.28) / 97.51(13.79) / \textbf{100.00(0.00)} & 95.17(2.27) / 98.71(1.98) / 98.35(9.25) / \textbf{100.00(0.00)} \\
    & & Uniform & 60.28(19.27) / 82.27(21.01) / 94.94(10.75) / \textbf{97.50(0.00)} & 88.72(7.18) / 94.66(7.88) / 97.42(13.85) / \textbf{100.00(0.00)} & 93.28(4.22) / 96.61(5.07) / 98.20(9.49) / \textbf{100.00(0.00)} \\
    \cmidrule(l){2-6} 
    & \multirow{13}{*}{\rotatebox[origin=c]{90}{CIFAR-100}} & 
        TIN (c)* & 60.26(1.93) / 70.35(5.42) / 65.48(12.80) / \textbf{89.81(1.52)} & 79.32(4.14) / 86.48(4.87) / 71.81(23.34) / \textbf{97.31(0.45)} & 80.66(6.76) / 85.96(6.62) / 73.56(20.11) / \textbf{97.55(0.37)} \\
    & & TIN (r) & 58.66(2.35) / 70.94(6.67) / 70.66(16.66) / \textbf{91.37(1.59)} & 77.07(6.35) / 86.12(6.70) / 75.25(25.95) / \textbf{97.82(0.53)} & 78.60(9.30) / 85.48(8.86) / 77.31(21.32) / \textbf{97.99(0.45)} \\
    & & LSUN (c)* & 58.61(0.57) / 67.31(3.92) / 59.41(7.88) / \textbf{85.26(0.79)} & 78.46(0.91) / 84.87(2.21) / 63.89(22.87) / \textbf{95.52(0.32)} & 80.86(0.93) / 85.02(1.94) / 66.59(20.06) / \textbf{95.82(0.29)} \\
    & & LSUN (r) & 59.32(2.47) / 72.23(7.36) / 70.16(17.28) / \textbf{90.69(2.23)} & 78.44(5.41) / 87.80(5.81) / 76.61(25.13) / \textbf{97.59(0.75)} & 80.74(6.55) / 87.99(6.59) / 79.78(19.97) / \textbf{97.83(0.65)} \\
    & & iSUN & 58.37(2.76) / 70.66(7.07) / 69.86(17.11) / \textbf{90.48(2.17)} & 76.89(6.28) / 86.28(6.40) / 75.20(25.92) / \textbf{97.45(0.73)} & 80.56(7.56) / 87.43(6.90) / 79.64(19.95) / \textbf{97.88(0.56)} \\
    & & SVNH & 56.40(1.82) / 60.28(8.31) / 64.99(14.82) / \textbf{88.42(2.57)} & 77.36(2.83) / 80.18(5.87) / 68.85(29.66) / \textbf{96.90(0.79)} & 66.58(4.07) / 67.51(7.64) / 60.65(31.79) / \textbf{93.95(1.33)} \\
    & & Food-101 & 62.64(0.63) / 61.81(5.88) / 50.09(2.38) / \textbf{68.29(1.18)} & 84.38(0.48) / 83.91(5.60) / 62.73(10.63) / \textbf{90.79(0.49)} & 77.21(0.77) / 74.57(8.77) / 48.80(12.33) / \textbf{86.96(0.59)} \\
    & & STL-10 & 57.82(1.32) / 67.22(4.44) / 64.71(13.15) / \textbf{75.20(3.96)} & 76.53(2.23) / 83.14(3.86) / 70.19(26.16) / \textbf{90.28(1.95)} & 81.56(2.67) / 85.23(3.91) / 76.73(20.01) / \textbf{92.43(1.54)} \\
    & & MNIST & 62.68(2.68) / \textbf{87.24(6.78)} / 64.53(20.93) / 86.62(5.51) & 81.56(3.83) / 96.09(3.00) / 83.64(22.90) / \textbf{96.17(2.06)} & 84.50(3.62) / 96.49(2.65) / 89.05(16.34) / \textbf{96.62(1.82)} \\
    & & F-MNIST & 70.15(1.95) / 90.24(3.60) / 62.67(18.30) / \textbf{94.69(1.10)} & 88.52(1.04) / 97.33(1.27) / 79.12(19.24) / \textbf{98.92(0.35)} & 90.30(0.80) / 97.54(1.07) / 84.07(15.08) / \textbf{99.02(0.31)} \\
    & & NotMNIST & 59.04(1.67) / 79.64(6.86) / 71.91(16.07) / \textbf{84.44(3.09)} & 79.97(1.37) / 92.35(3.38) / 82.46(22.02) / \textbf{95.80(0.86)} & 73.28(2.54) / 87.56(4.71) / 78.13(22.69) / \textbf{93.87(1.12)} \\
    & & Gaussian & 47.54(0.05) / 55.72(14.42) / 75.98(24.43) / \textbf{97.50(0.00)} & 53.73(13.50) / 61.56(31.80) / 68.81(40.93) / \textbf{99.74(0.52)} & 69.06(10.42) / 73.30(23.12) / 78.59(28.01) / \textbf{99.84(0.30)} \\
    & & Uniform & 48.34(1.81) / 60.07(20.04) / 77.50(23.34) / \textbf{97.50(0.01)} & 63.59(20.05) / 75.66(20.60) / 73.64(39.00) / \textbf{99.71(0.56)} & 75.90(13.16) / 83.94(14.22) / 81.48(26.97) / \textbf{99.82(0.34)} \\
    \midrule
    \multirow{24}{*}{\rotatebox[origin=c]{90}{WRN-28-10}} & \multirow{12}{*}{\rotatebox[origin=c]{90}{CIFAR-10}} & 
        TIN (c)* & 76.10(1.38) / 81.36(3.70) / 89.67(7.11) / \textbf{92.63(1.26)} & 91.79(1.57) / 90.35(3.81) / 96.50(4.12) / \textbf{98.17(0.33)} & 90.94(2.91) / 87.44(5.17) / 96.28(5.14) / \textbf{98.42(0.27)} \\
    & & TIN (r) & 72.35(2.13) / 79.39(4.11) / \textbf{91.47(7.96)} / 91.05(2.24) & 89.21(2.65) / 89.43(3.78) / 96.96(4.98) / \textbf{97.65(0.66)} & 87.62(4.65) / 86.83(5.01) / 96.60(6.01) / \textbf{97.94(0.54)} \\
    & & LSUN (c)* & 78.31(2.05) / 81.76(2.78) / 85.97(2.24) / \textbf{94.96(0.16)} & 93.67(0.50) / 90.12(3.69) / 95.69(1.16) / \textbf{98.98(0.07)} & 93.90(0.56) / 86.47(5.93) / 95.98(1.79) / \textbf{99.06(0.05)} \\
    & & LSUN (r) & 76.39(2.41) / 84.92(3.58) / 92.85(7.86) / \textbf{94.02(1.29)} & 92.45(1.48) / 93.48(2.45) / 97.78(4.34) / \textbf{98.59(0.34)} & 92.33(2.31) / 92.19(3.13) / 97.94(4.23) / \textbf{98.79(0.28)} \\
    & & iSUN & 74.82(2.19) / 83.11(3.82) / 92.54(7.77) / \textbf{93.47(1.35)} & 91.22(2.05) / 92.21(2.94) / 97.64(4.37) / \textbf{98.48(0.36)} & 91.40(3.31) / 91.44(3.60) / 97.89(4.47) / \textbf{98.82(0.26)} \\
    & & SVNH & 80.66(2.52) / 81.60(5.57) / 94.14(1.97) / \textbf{96.50(0.69)} & 94.43(1.30) / 90.52(4.39) / 98.63(0.83) / \textbf{99.52(0.24)} & 88.55(4.75) / 74.03(11.15) / 97.11(2.52) / \textbf{98.94(0.52)} \\
    & & Food-101 & 71.61(1.04) / 70.21(5.85) / 63.31(5.20) / \textbf{78.73(1.42)} & 89.71(0.90) / 80.29(8.31) / 84.19(5.85) / \textbf{93.95(0.41)} & 76.04(3.29) / 55.33(13.49) / 73.64(9.95) / \textbf{90.41(0.47)} \\
    & & MNIST & 72.35(5.94) / 87.91(6.39) / \textbf{95.08(4.88)} / 94.60(1.22) & 91.16(3.19) / 95.67(2.84) / \textbf{98.83(1.22)} / 98.60(0.39) & 91.37(3.95) / 95.22(3.27) / \textbf{99.19(0.86)} / 98.91(0.28) \\
    & & F-MNIST & 81.20(1.34) / 90.65(2.12) / 92.05(5.25) / \textbf{94.03(0.72)} & 94.59(0.61) / 96.71(0.96) / 97.73(1.85) / \textbf{98.60(0.24)} & 94.91(0.79) / 96.13(1.20) / 98.21(1.85) / \textbf{98.78(0.20)} \\
    & & NotMNIST & 81.27(2.34) / 92.38(3.23) / 95.95(2.67) / \textbf{96.33(0.62)} & 95.00(0.58) / 97.60(1.06) / \textbf{99.28(0.85)} / 99.11(0.28) & 92.37(1.43) / 94.82(1.96) / \textbf{98.85(1.40)} / 98.84(0.32) \\
    & & Gaussian & 82.94(17.44) / 93.87(7.97) / 97.12(2.80) / \textbf{97.50(0.00)} & 95.16(6.00) / 98.52(2.71) / 99.60(2.91) / \textbf{99.98(0.02)} & 96.47(4.70) / 98.80(2.29) / 99.39(4.51) / \textbf{99.99(0.01)} \\
    & & Uniform & 78.59(14.28) / 93.38(5.71) / 96.25(6.97) / \textbf{97.50(0.00)} & 95.30(2.90) / 98.52(1.69) / 98.01(13.34) / \textbf{99.99(0.01)} & 96.70(2.10) / 98.81(1.38) / 98.47(9.15) / \textbf{99.99(0.01)} \\
    \cmidrule(l){2-6} 
    & \multirow{13}{*}{\rotatebox[origin=c]{90}{CIFAR-100}} & 
        TIN (c)* & 61.96(0.78) / 69.18(2.82) / 77.40(12.03) / \textbf{85.01(1.35)} & 80.90(0.90) / 85.73(1.93) / 87.95(13.71) / \textbf{95.91(0.42)} & 81.38(1.79) / 85.44(2.16) / 88.31(13.86) / \textbf{96.48(0.31)} \\
    & & TIN (r) & 58.89(0.99) / 68.94(4.03) / 80.24(14.04) / \textbf{84.94(2.40)} & 76.67(2.03) / 84.60(3.21) / 88.31(16.07) / \textbf{95.84(0.67)} & 76.74(3.62) / 84.20(3.69) / 88.70(15.65) / \textbf{96.40(0.47)} \\
    & & LSUN (c)* & 58.99(0.77) / 66.46(5.62) / 69.79(7.98) / \textbf{82.97(1.77)} & 79.17(1.25) / 83.37(3.97) / 83.87(11.80) / \textbf{94.92(0.65)} & 80.88(1.99) / 82.91(4.16) / 84.43(12.89) / \textbf{95.45(0.57)} \\
    & & LSUN (r) & 58.98(1.47) / 69.64(4.44) / 79.14(14.57) / \textbf{82.76(2.40)} & 78.00(1.95) / 86.05(2.90) / 88.30(15.18) / \textbf{95.18(0.86)} & 79.25(2.28) / 86.09(2.87) / 88.98(14.72) / \textbf{95.92(0.68)} \\
    & & iSUN & 58.54(1.43) / 68.94(4.43) / 79.29(14.03) / \textbf{83.73(1.92)} & 77.29(2.15) / 85.38(3.03) / 88.01(15.49) / \textbf{95.39(0.55)} & 80.27(2.59) / 86.73(2.80) / 89.40(14.30) / \textbf{96.40(0.35)} \\
    & & SVNH & 58.13(2.90) / 66.73(10.20) / 76.30(14.18) / \textbf{90.28(1.57)} & 79.82(2.49) / 83.05(7.38) / 86.67(16.70) / \textbf{97.52(0.41)} & 66.44(4.78) / 68.06(11.64) / 79.01(22.03) / \textbf{95.13(0.54)} \\
    & & Food-101 & 69.12(0.76) / 63.20(4.76) / 58.46(6.74) / \textbf{72.68(1.01)} & 89.25(0.40) / 83.76(4.39) / 80.49(11.90) / \textbf{92.53(0.38)} & 83.20(0.70) / 70.63(8.14) / 73.20(16.92) / \textbf{89.37(0.54)} \\
    & & STL-10 & 58.23(0.55) / 64.30(2.64) / \textbf{72.25(11.96)} / 71.03(1.17) & 77.68(0.61) / 81.20(2.27) / 84.18(16.28) / \textbf{88.89(0.51)} & 81.66(1.01) / 83.70(2.31) / 87.95(12.62) / \textbf{91.70(0.43)} \\
    & & MNIST & 61.79(5.62) / 83.33(7.90) / 72.71(13.36) / \textbf{85.60(3.96)} & 82.64(4.73) / 95.07(3.39) / 87.62(12.89) / \textbf{96.03(1.49)} & 85.99(3.87) / 95.84(2.82) / 89.95(11.51) / \textbf{96.59(1.25)} \\
    & & F-MNIST & 72.39(2.13) / 86.47(3.21) / 76.60(10.40) / \textbf{92.48(1.22)} & 90.33(1.02) / 95.93(1.09) / 90.71(10.09) / \textbf{98.27(0.36)} & 91.74(0.87) / 96.18(0.97) / 92.33(8.75) / \textbf{98.46(0.31)} \\
    & & NotMNIST & 56.15(1.61) / 76.13(7.89) / \textbf{82.57(14.47)} / 77.23(2.62) & 80.00(2.79) / 91.23(3.99) / 89.53(17.24) / \textbf{93.05(1.47)} & 73.99(4.65) / 86.76(5.07) / 85.80(21.25) / \textbf{90.53(2.00)} \\
    & & Gaussian & 47.54(0.07) / 75.96(21.36) / 88.34(18.59) / \textbf{97.50(0.00)} & 59.59(30.02) / 82.89(24.64) / 88.04(30.26) / \textbf{99.82(0.15)} & 73.02(23.16) / 88.04(17.98) / 91.67(20.57) / \textbf{99.88(0.09)} \\
    \bottomrule
    \end{tabular}}
    \end{sc}
    \end{small}
    \end{center}
    \end{table}

    \clearpage
    In the above experiments, we used TIN(c)* and LSUN(c)*, i.e., our corrected version of the cropped images from TinyImageNet and LSUN; see Sec.~\ref{sec:black-frame-explain}. For the sake of completeness, we also conducted the same experiments using TIN(c) and LSUN(c), their original versions having a two-pixel black frame, supplied by the GitHub repo of the authors of ODIN [25]. Table \ref{table:less_biased_orig_cropped} shows the results.

    \begin{table}[h]
    \caption{More detailed results of OOD detection performance measured by the less-biased evaluation. The original cropped images are used instead of their `*' version.}
    \label{table:less_biased_orig_cropped}
    \begin{center}
    \begin{small}
    \begin{sc}
    \resizebox{\textwidth}{!}{
    \begin{tabular}{cccccc}
    \toprule
    Network & ID & OOD & Accuracy at TPR$=95\%$ & AUROC & AUPR-In  \\ \midrule
    & & & \multicolumn{3}{c}{Baseline\cite{hendrycks2016baseline}  /  ODIN\cite{liang2017enhancing}  /  Mahalanobis\cite{lee2018simple}  /  Ours} \\ \midrule
    \multirow{24}{*}{\rotatebox[origin=c]{90}{DenseBC}} & \multirow{12}{*}{\rotatebox[origin=c]{90}{CIFAR-10}} & 
        TIN (c) & 78.86(1.37) / 93.31(2.04) / 75.77(7.50) / \textbf{94.83(0.68)} & 94.90(0.43) / 98.31(0.82) / 86.44(8.15) / \textbf{98.89(0.24)} & 96.09(0.36) / 98.39(0.90) / 85.19(9.46) / \textbf{98.97(0.20)} \\
    & & TIN (r) & 73.51(2.48) / 88.47(3.99) / 82.87(13.81) / \textbf{94.67(0.60)} & 92.67(1.23) / 96.44(1.63) / 90.45(11.13) / \textbf{98.82(0.29)} & 94.06(1.17) / 96.42(1.92) / 89.51(12.38) / \textbf{98.89(0.26)} \\
    & & LSUN (c) & 81.22(1.46) / 92.22(1.49) / 65.60(6.43) / \textbf{95.27(0.19)} & 95.57(0.20) / 97.78(0.77) / 76.89(9.64) / \textbf{99.09(0.12)} & 96.59(0.11) / 97.62(1.11) / 75.44(10.64) / \textbf{99.12(0.11)} \\
    & & LSUN (r) & 76.74(1.40) / 92.09(2.51) / 83.55(13.63) / \textbf{95.72(0.56)} & 94.28(0.52) / 97.87(0.96) / 92.18(8.81) / \textbf{99.19(0.22)} & 95.60(0.44) / 97.99(1.00) / 92.19(9.12) / \textbf{99.26(0.20)} \\
    & & iSUN & 75.12(2.08) / 90.64(2.95) / 82.53(14.57) / \textbf{95.62(0.51)} & 93.62(0.83) / 97.34(1.14) / 90.59(11.06) / \textbf{99.20(0.19)} & 95.48(0.66) / 97.69(1.14) / 91.08(10.81) / \textbf{99.33(0.16)} \\
    & & SVNH & 68.84(2.90) / 76.56(8.01) / 74.45(20.13) / \textbf{95.36(0.87)} & 90.28(2.47) / 89.62(5.30) / 72.18(32.66) / \textbf{99.11(0.36)} & 82.90(7.23) / 75.67(11.31) / 65.64(35.76) / \textbf{97.99(0.73)} \\
    & & Food-101 & 67.93(0.59) / 71.93(5.11) / 56.95(4.79) / \textbf{79.79(1.03)} & 89.87(0.44) / 87.48(4.91) / 73.17(8.51) / \textbf{93.98(0.54)} & 83.58(0.96) / 73.56(11.24) / 57.03(11.25) / \textbf{89.99(0.90)} \\
    & & MNIST & 76.35(3.52) / \textbf{95.86(2.06)} / 91.42(11.63) / 95.15(1.51) & 94.54(1.02) / \textbf{99.26(0.67)} / 96.59(7.93) / 98.90(0.49) & 95.98(0.83) / \textbf{99.37(0.53)} / 97.60(5.92) / 99.06(0.42) \\
    & & F-MNIST & 81.65(2.11) / \textbf{95.72(1.25)} / 79.24(10.83) / 94.83(0.71) & 95.76(0.50) / \textbf{99.20(0.47)} / 92.32(7.38) / 98.84(0.21) & 96.76(0.35) / \textbf{99.29(0.39)} / 93.38(7.38) / 98.94(0.17) \\
    & & NotMNIST & 75.95(3.71) / 90.85(5.01) / 88.61(9.04) / \textbf{95.75(1.09)} & 93.72(1.49) / 97.33(1.72) / 94.06(8.89) / \textbf{99.01(0.30)} & 91.90(2.37) / 95.24(2.86) / 88.25(14.73) / \textbf{98.62(0.45)} \\
    & & Gaussian & 63.72(14.12) / 90.36(15.81) / 93.22(13.48) / \textbf{97.50(0.00)} & 91.96(4.29) / 97.95(3.19) / 96.12(15.44) / \textbf{100.00(0.00)} & 95.17(2.27) / 98.73(1.92) / 97.43(10.34) / \textbf{100.00(0.00)} \\
    & & Uniform & 60.28(19.27) / 81.82(21.23) / 92.92(13.69) / \textbf{97.50(0.00)} & 88.72(7.18) / 94.63(7.73) / 96.62(14.15) / \textbf{100.00(0.00)} & 93.28(4.22) / 96.59(4.96) / 97.67(9.68) / \textbf{100.00(0.00)} \\
    \cmidrule(l){2-6} 
    & \multirow{13}{*}{\rotatebox[origin=c]{90}{CIFAR-100}} & 
        TIN (c) & 64.67(2.28) / 81.15(6.13) / 66.07(10.16) / \textbf{91.71(0.85)} & 83.70(4.00) / 92.79(3.90) / 73.92(19.34) / \textbf{97.90(0.29)} & 84.98(6.13) / 92.52(5.05) / 73.78(15.87) / \textbf{98.03(0.24)} \\
    & & TIN (r) & 58.66(2.35) / 71.32(6.55) / 68.00(16.07) / \textbf{91.37(1.59)} & 77.07(6.35) / 86.35(6.63) / 71.75(25.16) / \textbf{97.82(0.53)} & 78.60(9.30) / 85.68(8.80) / 73.14(20.57) / \textbf{97.99(0.45)} \\
    & & LSUN (c) & 62.53(0.41) / 79.02(3.86) / 60.97(11.23) / \textbf{88.48(0.77)} & 82.92(0.59) / 92.58(1.87) / 66.52(19.93) / \textbf{96.73(0.31)} & 85.41(0.77) / 92.98(1.75) / 67.46(18.44) / \textbf{96.89(0.28)} \\
    & & LSUN (r) & 59.32(2.47) / 72.75(7.20) / 67.95(16.61) / \textbf{90.69(2.23)} & 78.44(5.41) / 88.13(5.65) / 73.90(24.51) / \textbf{97.59(0.75)} & 80.74(6.55) / 88.29(6.42) / 76.66(19.51) / \textbf{97.83(0.65)} \\
    & & iSUN & 58.37(2.76) / 71.08(6.98) / 67.10(16.53) / \textbf{90.48(2.17)} & 76.89(6.28) / 86.56(6.29) / 71.71(25.16) / \textbf{97.45(0.73)} & 80.56(7.56) / 87.67(6.78) / 76.00(19.33) / \textbf{97.88(0.56)} \\
    & & SVNH & 56.40(1.82) / 59.66(7.63) / 60.25(13.81) / \textbf{88.42(2.57)} & 77.36(2.83) / 79.74(5.50) / 61.99(27.81) / \textbf{96.90(0.79)} & 66.58(4.07) / 66.79(7.23) / 50.78(29.99) / \textbf{93.95(1.33)} \\
    & & Food-101 & 62.64(0.63) / 62.29(5.44) / 50.50(2.39) / \textbf{68.29(1.18)} & 84.38(0.48) / 84.41(5.15) / 61.97(10.18) / \textbf{90.79(0.49)} & 77.21(0.77) / 75.20(8.23) / 46.60(11.77) / \textbf{86.96(0.59)} \\
    & & STL-10 & 57.82(1.32) / 67.57(4.27) / 64.44(12.97) / \textbf{75.20(3.96)} & 76.53(2.23) / 83.43(3.67) / 67.71(25.77) / \textbf{90.28(1.95)} & 81.56(2.67) / 85.48(3.72) / 73.69(19.52) / \textbf{92.43(1.54)} \\
    & & MNIST & 62.68(2.68) / \textbf{87.43(6.85)} / 71.45(22.63) / 86.62(5.51) & 81.56(3.83) / 96.16(3.01) / 85.24(23.54) / \textbf{96.17(2.06)} & 84.50(3.62) / 96.55(2.66) / 89.93(16.70) / \textbf{96.62(1.82)} \\
    & & F-MNIST & 70.15(1.95) / 90.46(3.67) / 67.74(19.52) / \textbf{94.69(1.10)} & 88.52(1.04) / 97.41(1.28) / 80.36(20.33) / \textbf{98.92(0.35)} & 90.30(0.80) / 97.61(1.09) / 84.05(16.58) / \textbf{99.02(0.31)} \\
    & & NotMNIST & 59.04(1.67) / 79.54(6.79) / 75.28(16.97) / \textbf{84.44(3.09)} & 79.97(1.37) / 92.29(3.37) / 83.24(22.35) / \textbf{95.80(0.86)} & 73.28(2.54) / 87.43(4.71) / 78.33(22.92) / \textbf{93.87(1.12)} \\
    & & Gaussian & 47.54(0.05) / 54.60(13.58) / 70.35(24.18) / \textbf{97.50(0.00)} & 53.73(13.50) / 59.75(31.22) / 62.78(41.05) / \textbf{99.74(0.52)} & 69.06(10.42) / 72.10(22.75) / 74.21(27.99) / \textbf{99.84(0.30)} \\
    & & Uniform & 48.34(1.81) / 59.13(19.58) / 72.29(23.29) / \textbf{97.50(0.01)} & 63.59(20.05) / 74.60(20.42) / 68.71(38.75) / \textbf{99.71(0.56)} & 75.90(13.16) / 83.24(14.11) / 77.56(27.12) / \textbf{99.82(0.34)} \\
    \midrule
    \multirow{24}{*}{\rotatebox[origin=c]{90}{WRN-28-10}} & \multirow{12}{*}{\rotatebox[origin=c]{90}{CIFAR-10}} & 
        TIN (c) & 78.71(1.31) / 86.55(2.87) / 78.05(7.26) / \textbf{93.26(0.96)} & 93.86(0.90) / 95.03(1.74) / 90.53(5.96) / \textbf{98.35(0.32)} & 94.45(1.23) / 94.50(2.05) / 90.09(6.58) / \textbf{98.53(0.27)} \\
    & & TIN (r) & 72.35(2.13) / 79.50(4.13) / 90.84(7.94) / \textbf{91.05(2.24)} & 89.21(2.65) / 89.43(3.78) / 96.62(4.95) / \textbf{97.65(0.66)} & 87.62(4.65) / 86.82(4.98) / 96.15(5.98) / \textbf{97.94(0.54)} \\
    & & LSUN (c) & 82.07(1.93) / 87.51(1.97) / 69.63(5.03) / \textbf{95.58(0.21)} & 95.41(0.26) / 94.92(1.72) / 85.47(5.73) / \textbf{99.19(0.07)} & 96.19(0.18) / 93.83(2.66) / 85.34(7.15) / \textbf{99.23(0.07)} \\
    & & LSUN (r) & 76.39(2.41) / 85.04(3.58) / 92.36(7.84) / \textbf{94.02(1.29)} & 92.45(1.48) / 93.49(2.45) / 97.58(4.32) / \textbf{98.59(0.34)} & 92.33(2.31) / 92.19(3.12) / 97.72(4.22) / \textbf{98.79(0.28)} \\
    & & iSUN & 74.82(2.19) / 83.22(3.82) / 91.92(7.76) / \textbf{93.47(1.35)} & 91.22(2.05) / 92.22(2.94) / 97.38(4.35) / \textbf{98.48(0.36)} & 91.40(3.31) / 91.44(3.58) / 97.62(4.45) / \textbf{98.82(0.26)} \\
    & & SVNH & 80.66(2.52) / 81.36(5.60) / 93.77(2.04) / \textbf{96.50(0.69)} & 94.43(1.30) / 90.31(4.41) / 98.47(0.86) / \textbf{99.52(0.24)} & 88.55(4.75) / 73.47(11.23) / 96.80(2.55) / \textbf{98.94(0.52)} \\
    & & Food-101 & 71.61(1.04) / 70.11(5.79) / 63.73(5.36) / \textbf{78.73(1.42)} & 89.71(0.90) / 80.07(8.16) / 84.27(5.84) / \textbf{93.95(0.41)} & 76.04(3.29) / 54.80(13.10) / 73.44(9.85) / \textbf{90.41(0.47)} \\
    & & MNIST & 72.35(5.94) / 88.24(5.90) / \textbf{94.88(5.09)} / 94.60(1.22) & 91.16(3.19) / 95.80(2.62) / \textbf{98.77(1.33)} / 98.60(0.39) & 91.37(3.95) / 95.36(3.02) / \textbf{99.13(0.98)} / 98.91(0.28) \\
    & & F-MNIST & 81.20(1.34) / 90.75(2.10) / 91.51(5.27) / \textbf{94.03(0.72)} & 94.59(0.61) / 96.73(0.96) / 97.54(1.86) / \textbf{98.60(0.24)} & 94.91(0.79) / 96.15(1.20) / 98.01(1.88) / \textbf{98.78(0.20)} \\
    & & NotMNIST & 81.27(2.34) / 92.55(3.10) / 95.61(2.86) / \textbf{96.33(0.62)} & 95.00(0.58) / 97.66(1.01) / 99.08(1.08) / \textbf{99.11(0.28)} & 92.37(1.43) / 94.90(1.86) / 98.39(2.27) / \textbf{98.84(0.32)} \\
    & & Gaussian & 82.94(17.44) / 94.05(7.44) / 97.12(2.80) / \textbf{97.50(0.00)} & 95.16(6.00) / 98.61(2.50) / 99.60(2.91) / \textbf{99.98(0.02)} & 96.47(4.70) / 98.87(2.11) / 99.39(4.51) / \textbf{99.99(0.01)} \\
    & & Uniform & 78.59(14.28) / 93.77(5.26) / 96.25(6.97) / \textbf{97.50(0.00)} & 95.30(2.90) / 98.61(1.58) / 98.01(13.34) / \textbf{99.99(0.01)} & 96.70(2.10) / 98.87(1.30) / 98.47(9.15) / \textbf{99.99(0.01)} \\
    \cmidrule(l){2-6} 
    & \multirow{13}{*}{\rotatebox[origin=c]{90}{CIFAR-100}} & 
        TIN (c) & 65.24(1.62) / 75.01(3.51) / 67.17(8.92) / \textbf{87.94(1.16)} & 84.47(1.24) / 90.67(1.73) / 81.80(11.04) / \textbf{96.76(0.34)} & 86.08(1.27) / 91.31(1.49) / 81.95(11.95) / \textbf{97.11(0.25)} \\
    & & TIN (r) & 58.89(0.99) / 68.95(3.98) / 79.33(13.77) / \textbf{84.94(2.40)} & 76.67(2.03) / 84.62(3.16) / 87.65(15.93) / \textbf{95.84(0.67)} & 76.74(3.62) / 84.22(3.63) / 87.92(15.56) / \textbf{96.40(0.47)} \\
    & & LSUN (c) & 60.99(1.01) / 71.10(5.89) / 59.21(5.80) / \textbf{86.49(1.74)} & 81.91(1.31) / 88.32(3.51) / 74.40(5.23) / \textbf{96.09(0.62)} & 84.68(1.45) / 89.14(3.00) / 74.50(5.80) / \textbf{96.41(0.56)} \\
    & & LSUN (r) & 58.98(1.47) / 69.67(4.41) / 78.24(14.33) / \textbf{82.76(2.40)} & 78.00(1.95) / 86.07(2.87) / 87.56(15.15) / \textbf{95.18(0.86)} & 79.25(2.28) / 86.11(2.84) / 88.12(14.79) / \textbf{95.92(0.68)} \\
    & & iSUN & 58.54(1.43) / 68.97(4.40) / 78.27(13.79) / \textbf{83.73(1.92)} & 77.29(2.15) / 85.40(3.01) / 87.20(15.42) / \textbf{95.39(0.55)} & 80.27(2.59) / 86.75(2.78) / 88.57(14.29) / \textbf{96.40(0.35)} \\
    & & SVNH & 58.13(2.90) / 66.89(10.04) / 73.78(15.04) / \textbf{90.28(1.57)} & 79.82(2.49) / 83.23(7.27) / 83.92(18.34) / \textbf{97.52(0.41)} & 66.44(4.78) / 68.31(11.47) / 74.78(24.30) / \textbf{95.13(0.54)} \\
    & & Food-101 & 69.12(0.76) / 63.17(4.72) / 58.03(6.90) / \textbf{72.68(1.01)} & 89.25(0.40) / 83.72(4.37) / 79.23(12.65) / \textbf{92.53(0.38)} & 83.20(0.70) / 70.54(8.15) / 71.23(18.04) / \textbf{89.37(0.54)} \\
    & & STL-10 & 58.23(0.55) / 64.29(2.56) / \textbf{71.78(11.79)} / 71.03(1.17) & 77.68(0.61) / 81.19(2.19) / 83.70(16.12) / \textbf{88.89(0.51)} & 81.66(1.01) / 83.70(2.23) / 87.45(12.48) / \textbf{91.70(0.43)} \\
    & & MNIST & 61.79(5.62) / 83.44(7.86) / 72.43(13.40) / \textbf{85.60(3.96)} & 82.64(4.73) / 95.11(3.39) / 87.26(12.91) / \textbf{96.03(1.49)} & 85.99(3.87) / 95.87(2.82) / 89.60(11.51) / \textbf{96.59(1.25)} \\
    & & F-MNIST & 72.39(2.13) / 86.51(3.21) / 75.80(10.71) / \textbf{92.48(1.22)} & 90.33(1.02) / 95.94(1.09) / 90.00(10.31) / \textbf{98.27(0.36)} & 91.74(0.87) / 96.19(0.97) / 91.63(8.96) / \textbf{98.46(0.31)} \\
    & & NotMNIST & 56.15(1.61) / 76.13(7.88) / \textbf{80.87(15.35)} / 77.23(2.62) & 80.00(2.79) / 91.24(4.00) / 87.56(18.87) / \textbf{93.05(1.47)} & 73.99(4.65) / 86.78(5.08) / 83.07(23.07) / \textbf{90.53(2.00)} \\
    & & Gaussian & 47.54(0.07) / 75.81(21.68) / 87.95(18.57) / \textbf{97.50(0.00)} & 59.59(30.02) / 83.00(24.41) / 87.92(30.23) / \textbf{99.82(0.15)} & 73.02(23.16) / 88.14(17.78) / 91.59(20.54) / \textbf{99.88(0.09)} \\
    & & Uniform & 47.51(0.02) / 68.50(20.65) / 88.96(17.14) / \textbf{97.50(0.00)} & 60.88(14.02) / 88.19(12.01) / 88.33(29.70) / \textbf{99.95(0.10)} & 74.03(12.70) / 92.15(8.43) / 91.62(20.71) / \textbf{99.97(0.07)} \\
    \bottomrule
    \end{tabular}}
    \end{sc}
    \end{small}
    \end{center}
    \end{table}
    
\clearpage
    
\section{Dependency on an OOD Validation Dataset: Full Version}%The OOD Detection Performance Tuned by OOD Validation dataset}

\begin{figure}[h]
    \centerline{\includegraphics[width=0.8\textwidth]{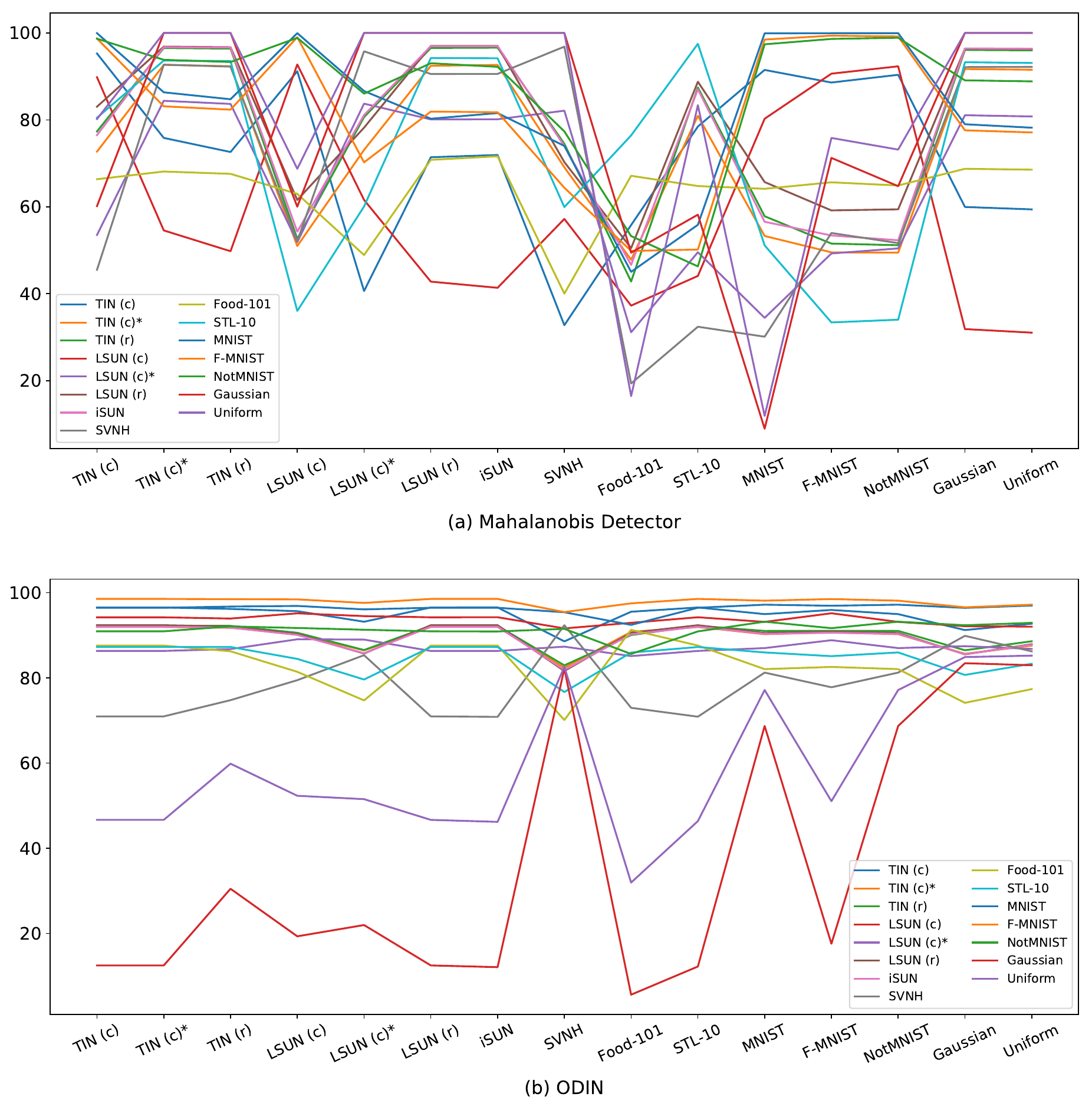}}
    \caption{Dependency of (a) Mahalanobis Detector and (b) ODIN on the assumed OOD dataset used for hyperparameter determination. AUROC of detecting OOD samples given in the horizontal axis. The line colors indicate the assumed OOD dataset. }
    \label{fig:tuning_performance_full}
    \end{figure}

    In Fig.~\ref{fig::Tuning_Maha} of the main paper, we demonstrate the dependency of the previous methods on the assumed OOD dataset used for hyperparameter determination, where only TinyImageNet (resized) and F-MNIST are used as the assumed datasets. We show here the complete results in Fig.~\ref{fig:tuning_performance_full}; it shows AUROC of detecting OOD samples given in the horizontal axis when CIFAR-100 is the ID dataset and WRN-28-10 is employed.    

\clearpage
    
\section{More Complete Results of One-vs-one Evaluation}

    Table \ref{table:OOD_detection_result_one_vs_one} shows a more complete version of Table \ref{table:single_ood_result2} in the main paper. It shows the performances measured by the two additional metrics as above.
    
    \begin{table}[h]
    \caption{Performance of four out-of-distribution detection methods on a single network using one-vs-one evaluation.}
    \label{table:OOD_detection_result_one_vs_one}
    \begin{center}
    \begin{small}
    \begin{sc}
    \resizebox{\textwidth}{!}{
    \begin{tabular}{cccccc}
    \toprule
    Network & ID & OOD & Accuracy at TPR$=95\%$ & AUROC & AUPR-In  \\ \midrule
    & & & \multicolumn{3}{c}{Baseline\cite{hendrycks2016baseline}  /  ODIN\cite{liang2017enhancing}  /  Mahalanobis\cite{lee2018simple}  /  Ours} \\ \midrule
    \multirow{28}{*}{\rotatebox[origin=c]{90}{DenseBC}} & \multirow{14}{*}{\rotatebox[origin=c]{90}{CIFAR-10}} & 
        TIN (c) & 78.86(1.37) / 94.59(0.80) / 85.51(2.28) / \textbf{94.83(0.68)} & 94.90(0.43) / 98.79(0.32) / 94.48(1.19) / \textbf{98.89(0.24)} & 96.09(0.36) / 98.86(0.28) / 93.76(1.67) / \textbf{98.97(0.20)} \\
    & & TIN (c)* & 74.53(1.93) / 88.97(2.54) / 90.84(0.96) / \textbf{94.34(0.61)} & 93.26(0.85) / 96.67(0.97) / 97.36(0.39) / \textbf{98.74(0.23)} & 94.61(0.80) / 96.64(0.96) / 97.64(0.37) / \textbf{98.86(0.19)} \\
    & & TIN (r) & 73.51(2.48) / 90.39(2.98) / \textbf{95.00(0.54)} / 94.67(0.60) & 92.67(1.23) / 97.20(1.17) / \textbf{98.91(0.23)} / 98.82(0.29) & 94.06(1.17) / 97.27(1.10) / \textbf{99.04(0.20)} / 98.89(0.26) \\
    & & LSUN (c) & 81.22(1.46) / 93.67(0.53) / 74.91(5.87) / \textbf{95.27(0.19)} & 95.57(0.20) / 98.48(0.14) / 89.06(3.21) / \textbf{99.09(0.12)} & 96.59(0.11) / 98.62(0.16) / 89.00(2.32) / \textbf{99.12(0.11)} \\
    & & LSUN (c)* & 75.35(1.90) / 87.61(1.40) / 82.51(1.62) / \textbf{94.54(0.39)} & 93.72(0.39) / 96.41(0.52) / 93.63(0.69) / \textbf{98.83(0.18)} & 95.06(0.27) / 96.69(0.53) / 94.21(0.58) / \textbf{98.88(0.17)} \\
    & & LSUN (r) & 76.74(1.40) / 93.63(1.37) / 95.48(0.43) / \textbf{95.72(0.56)} & 94.28(0.52) / 98.43(0.49) / 99.00(0.23) / \textbf{99.19(0.22)} & 95.60(0.44) / 98.52(0.44) / 99.15(0.20) / \textbf{99.26(0.20)} \\
    & & iSUN & 75.12(2.08) / 92.26(1.99) / 95.09(0.51) / \textbf{95.62(0.51)} & 93.62(0.83) / 97.92(0.71) / 98.95(0.21) / \textbf{99.20(0.19)} & 95.48(0.66) / 98.21(0.58) / 99.22(0.15) / \textbf{99.33(0.16)} \\
    & & SVNH & 68.84(2.90) / 88.83(0.28) / 95.01(0.78) / \textbf{95.36(0.87)} & 90.28(2.47) / 95.11(0.48) / 98.89(0.37) / \textbf{99.11(0.36)} & 82.90(7.23) / 83.30(1.91) / 97.67(0.73) / \textbf{97.99(0.73)} \\
    & & Food-101 & 67.93(0.59) / 76.90(1.39) / 64.14(1.46) / \textbf{79.79(1.03)} & 89.87(0.44) / 92.06(0.71) / 80.38(3.83) / \textbf{93.98(0.54)} & 83.58(0.96) / 85.30(1.54) / 63.50(8.01) / \textbf{89.99(0.90)} \\
    & & MNIST & 76.35(3.52) / 97.00(0.49) / \textbf{97.50(0.00)} / 95.15(1.51) & 94.54(1.02) / \textbf{99.69(0.20)} / 99.43(0.44) / 98.90(0.49) & 95.98(0.83) / \textbf{99.72(0.18)} / 99.68(0.25) / 99.06(0.42) \\
    & & F-MNIST & 81.65(2.11) / \textbf{96.49(0.39)} / 92.79(4.59) / 94.83(0.71) & 95.76(0.50) / \textbf{99.51(0.13)} / 97.87(2.00) / 98.84(0.21) & 96.76(0.35) / \textbf{99.54(0.11)} / 97.95(2.26) / 98.94(0.17) \\
    & & NotMNIST & 75.95(3.71) / 94.44(2.12) / \textbf{96.04(1.24)} / 95.75(1.09) & 93.72(1.49) / 98.61(0.95) / \textbf{99.18(0.60)} / 99.01(0.30) & 91.90(2.37) / 97.04(2.34) / \textbf{98.98(0.63)} / 98.62(0.45) \\
    & & Gaussian & 63.72(14.12) / 97.48(0.03) / \textbf{97.50(0.00)} / \textbf{97.50(0.00)} & 91.96(4.29) / 99.30(0.47) / \textbf{100.00(0.00)} / \textbf{100.00(0.00)} & 95.17(2.27) / 99.54(0.26) / \textbf{100.00(0.00)} / \textbf{100.00(0.00)} \\
    & & Uniform & 60.28(19.27) / 96.79(1.52) / \textbf{97.50(0.00)} / \textbf{97.50(0.00)} & 88.72(7.18) / 98.79(0.95) / \textbf{100.00(0.00)} / \textbf{100.00(0.00)} & 93.28(4.22) / 99.23(0.57) / \textbf{100.00(0.00)} / \textbf{100.00(0.00)} \\
    \cmidrule(l){2-6} 
    & \multirow{15}{*}{\rotatebox[origin=c]{90}{CIFAR-100}} & 
        TIN (c) & 64.67(2.28) / 84.84(4.02) / 81.53(4.16) / \textbf{91.71(0.85)} & 83.70(4.00) / 94.48(3.21) / 92.97(1.63) / \textbf{97.90(0.29)} & 84.98(6.13) / 94.16(4.45) / 93.00(1.86) / \textbf{98.03(0.24)} \\
    & & TIN (c)* & 60.26(1.93) / 73.44(4.06) / 80.13(0.82) / \textbf{89.81(1.52)} & 79.32(4.14) / 88.54(4.27) / 93.18(0.39) / \textbf{97.31(0.45)} & 80.66(6.76) / 87.97(6.39) / 94.05(0.42) / \textbf{97.55(0.37)} \\
    & & TIN (r) & 58.66(2.35) / 74.66(5.82) / 88.95(0.58) / \textbf{91.37(1.59)} & 77.07(6.35) / 88.14(6.92) / 96.81(0.27) / \textbf{97.82(0.53)} & 78.60(9.30) / 87.18(9.82) / 97.26(0.30) / \textbf{97.99(0.45)} \\
    & & LSUN (c) & 62.53(0.41) / 84.21(1.09) / 77.94(5.23) / \textbf{88.48(0.77)} & 82.92(0.59) / 94.72(0.59) / 91.65(2.96) / \textbf{96.73(0.31)} & 85.41(0.77) / 94.80(0.69) / 91.67(2.98) / \textbf{96.89(0.28)} \\
    & & LSUN (c)* & 58.61(0.57) / 73.30(1.74) / 70.53(0.95) / \textbf{85.26(0.79)} & 78.46(0.91) / 87.89(1.13) / 85.44(1.85) / \textbf{95.52(0.32)} & 80.86(0.93) / 87.55(1.35) / 86.39(2.20) / \textbf{95.82(0.29)} \\
    & & LSUN (r) & 59.32(2.47) / 76.53(5.59) / 89.82(0.48) / \textbf{90.69(2.23)} & 78.44(5.41) / 90.38(4.76) / 97.00(0.15) / \textbf{97.59(0.75)} & 80.74(6.55) / 90.49(5.57) / 97.54(0.10) / \textbf{97.83(0.65)} \\
    & & iSUN & 58.37(2.76) / 74.54(6.13) / 89.43(0.20) / \textbf{90.48(2.17)} & 76.89(6.28) / 88.27(6.49) / 97.04(0.10) / \textbf{97.45(0.73)} & 80.56(7.56) / 89.01(7.25) / 97.80(0.13) / \textbf{97.88(0.56)} \\
    & & SVNH & 56.40(1.82) / 78.49(1.49) / 87.06(2.51) / \textbf{88.42(2.57)} & 77.36(2.83) / 91.60(0.73) / 96.48(0.68) / \textbf{96.90(0.79)} & 66.58(4.07) / 82.08(1.58) / \textbf{94.33(0.78)} / 93.95(1.33) \\
    & & Food-101 & 62.64(0.63) / \textbf{70.96(1.24)} / 55.43(2.48) / 68.29(1.18) & 84.38(0.48) / \textbf{90.82(0.60)} / 67.14(1.39) / 90.79(0.49) & 77.21(0.77) / 85.64(1.04) / 46.59(4.01) / \textbf{86.96(0.59)} \\
    & & STL-10 & 57.82(1.32) / 70.35(2.12) / \textbf{85.77(6.70)} / 75.20(3.96) & 76.53(2.23) / 85.41(2.55) / \textbf{91.64(5.03)} / 90.28(1.95) & 81.56(2.67) / 87.27(2.78) / 91.32(5.15) / \textbf{92.43(1.54)} \\
    & & MNIST & 62.68(2.68) / 91.17(5.19) / \textbf{97.47(0.07)} / 86.62(5.51) & 81.56(3.83) / 97.57(2.07) / \textbf{99.81(0.19)} / 96.17(2.06) & 84.50(3.62) / 97.74(1.94) / \textbf{99.88(0.12)} / 96.62(1.82) \\
    & & F-MNIST & 70.15(1.95) / 93.01(1.50) / \textbf{94.79(2.94)} / 94.69(1.10) & 88.52(1.04) / 98.27(0.49) / 98.58(1.11) / \textbf{98.92(0.35)} & 90.30(0.80) / 98.33(0.48) / 98.66(1.18) / \textbf{99.02(0.31)} \\
    & & NotMNIST & 59.04(1.67) / 85.20(2.34) / \textbf{94.73(2.98)} / 84.44(3.09) & 79.97(1.37) / 94.73(1.37) / \textbf{98.69(1.02)} / 95.80(0.86) & 73.28(2.54) / 90.74(2.83) / \textbf{97.98(0.88)} / 93.87(1.12) \\
    & & Gaussian & 47.54(0.05) / 77.72(19.57) / \textbf{97.50(0.00)} / \textbf{97.50(0.00)} & 53.73(13.50) / 93.44(6.60) / \textbf{100.00(0.00)} / 99.74(0.52) & 69.06(10.42) / 95.41(4.73) / \textbf{100.00(0.00)} / 99.84(0.30) \\
    & & Uniform & 48.34(1.81) / 79.64(23.92) / \textbf{97.50(0.00)} / \textbf{97.50(0.00)} & 63.59(20.05) / 94.60(6.66) / \textbf{100.00(0.00)} / 99.71(0.56) & 75.90(13.16) / 96.41(4.48) / \textbf{100.00(0.00)} / 99.82(0.34) \\
    \midrule
    \multirow{28}{*}{\rotatebox[origin=c]{90}{WRN-28-10}} & \multirow{14}{*}{\rotatebox[origin=c]{90}{CIFAR-10}} & 
        TIN (c) & 78.71(1.31) / 88.38(1.69) / 86.46(2.18) / \textbf{93.26(0.96)} & 93.86(0.90) / 95.88(1.01) / 95.99(1.04) / \textbf{98.35(0.32)} & 94.45(1.23) / 95.38(1.23) / 96.55(0.98) / \textbf{98.53(0.27)} \\
    & & TIN (c)* & 76.10(1.38) / 83.16(2.07) / \textbf{93.57(0.31)} / 92.63(1.26) & 91.79(1.57) / 92.17(2.19) / \textbf{98.50(0.11)} / 98.17(0.33) & 90.94(2.91) / 89.77(3.56) / \textbf{98.72(0.10)} / 98.42(0.27) \\
    & & TIN (r) & 72.35(2.13) / 81.86(3.29) / \textbf{95.59(0.34)} / 91.05(2.24) & 89.21(2.65) / 90.60(3.21) / \textbf{99.15(0.18)} / 97.65(0.66) & 87.62(4.65) / 87.99(4.53) / \textbf{99.25(0.19)} / 97.94(0.54) \\
    & & LSUN (c) & 82.07(1.93) / 90.24(0.84) / 78.82(2.25) / \textbf{95.58(0.21)} & 95.41(0.26) / 97.20(0.15) / 92.65(1.33) / \textbf{99.19(0.07)} & 96.19(0.18) / 97.28(0.17) / 93.12(1.38) / \textbf{99.23(0.07)} \\
    & & LSUN (c)* & 78.31(2.05) / 85.89(1.53) / 88.26(1.26) / \textbf{94.96(0.16)} & 93.67(0.50) / 95.08(0.42) / 96.90(0.35) / \textbf{98.98(0.07)} & 93.90(0.56) / 94.42(0.56) / 97.46(0.27) / \textbf{99.06(0.05)} \\
    & & LSUN (r) & 76.39(2.41) / 87.33(2.22) / \textbf{96.40(0.31)} / 94.02(1.29) & 92.45(1.48) / 94.48(1.70) / \textbf{99.37(0.13)} / 98.59(0.34) & 92.33(2.31) / 93.12(2.42) / \textbf{99.47(0.10)} / 98.79(0.28) \\
    & & iSUN & 74.82(2.19) / 85.49(2.99) / \textbf{96.08(0.20)} / 93.47(1.35) & 91.22(2.05) / 93.25(2.43) / \textbf{99.29(0.10)} / 98.48(0.36) & 91.40(3.31) / 92.35(3.24) / \textbf{99.46(0.08)} / 98.82(0.26) \\
    & & SVNH & 80.66(2.52) / 87.37(3.32) / 95.79(0.22) / \textbf{96.50(0.69)} & 94.43(1.30) / 93.34(3.60) / 99.28(0.09) / \textbf{99.52(0.24)} & 88.55(4.75) / 79.63(11.13) / 98.51(0.16) / \textbf{98.94(0.52)} \\
    & & Food-101 & 71.61(1.04) / 75.92(1.90) / 71.73(1.64) / \textbf{78.73(1.42)} & 89.71(0.90) / 89.18(2.37) / 90.43(1.54) / \textbf{93.95(0.41)} & 76.04(3.29) / 71.88(6.62) / 84.07(3.15) / \textbf{90.41(0.47)} \\
    & & MNIST & 72.35(5.94) / 92.32(2.53) / \textbf{97.42(0.05)} / 94.60(1.22) & 91.16(3.19) / 97.36(1.50) / \textbf{99.56(0.11)} / 98.60(0.39) & 91.37(3.95) / 97.03(1.79) / \textbf{99.71(0.07)} / 98.91(0.28) \\
    & & F-MNIST & 81.20(1.34) / 92.25(1.25) / \textbf{96.24(0.68)} / 94.03(0.72) & 94.59(0.61) / 97.28(0.65) / \textbf{99.03(0.23)} / 98.60(0.24) & 94.91(0.79) / 96.65(0.91) / \textbf{99.28(0.21)} / 98.78(0.20) \\
    & & NotMNIST & 81.27(2.34) / 94.93(0.51) / \textbf{97.33(0.14)} / 96.33(0.62) & 95.00(0.58) / 98.48(0.38) / \textbf{99.77(0.07)} / 99.11(0.28) & 92.37(1.43) / 96.20(1.28) / \textbf{99.65(0.10)} / 98.84(0.32) \\
    & & Gaussian & 82.94(17.44) / 96.33(2.54) / \textbf{97.50(0.00)} / \textbf{97.50(0.00)} & 95.16(6.00) / 99.39(0.94) / \textbf{100.00(0.00)} / 99.98(0.02) & 96.47(4.70) / 99.48(0.82) / \textbf{100.00(0.00)} / 99.99(0.01) \\
    & & Uniform & 78.59(14.28) / 96.65(1.53) / \textbf{97.50(0.00)} / \textbf{97.50(0.00)} & 95.30(2.90) / 99.41(0.53) / \textbf{100.00(0.00)} / 99.99(0.01) & 96.70(2.10) / 99.48(0.53) / \textbf{100.00(0.00)} / 99.99(0.01) \\
    \cmidrule(l){2-6} 
    & \multirow{15}{*}{\rotatebox[origin=c]{90}{CIFAR-100}} & 
        TIN (c) & 65.24(1.62) / 77.12(2.53) / 79.69(5.23) / \textbf{87.94(1.16)} & 84.47(1.24) / 91.72(1.10) / 92.58(2.60) / \textbf{96.76(0.34)} & 86.08(1.27) / 92.21(0.95) / 93.37(2.43) / \textbf{97.11(0.25)} \\
    & & TIN (c)* & 61.96(0.78) / 71.32(1.31) / \textbf{87.88(0.94)} / 85.01(1.35) & 80.90(0.90) / 87.08(1.29) / \textbf{96.45(0.30)} / 95.91(0.42) & 81.38(1.79) / 86.72(1.71) / \textbf{97.02(0.29)} / 96.48(0.31) \\
    & & TIN (r) & 58.89(0.99) / 71.53(2.26) / \textbf{91.88(0.30)} / 84.94(2.40) & 76.67(2.03) / 86.28(2.43) / \textbf{97.82(0.13)} / 95.84(0.67) & 76.74(3.62) / 85.82(3.03) / \textbf{98.09(0.10)} / 96.40(0.47) \\
    & & LSUN (c) & 60.99(1.01) / 78.04(0.67) / 68.64(0.78) / \textbf{86.49(1.74)} & 81.91(1.31) / 91.75(0.44) / 80.48(1.14) / \textbf{96.09(0.62)} & 84.68(1.45) / 91.90(0.47) / 79.46(1.59) / \textbf{96.41(0.56)} \\
    & & LSUN (c)* & 58.99(0.77) / 74.15(0.57) / 77.82(1.07) / \textbf{82.97(1.77)} & 79.17(1.25) / 88.06(0.46) / 91.13(0.52) / \textbf{94.92(0.65)} & 80.88(1.99) / 87.28(0.65) / 92.04(0.52) / \textbf{95.45(0.57)} \\
    & & LSUN (r) & 58.98(1.47) / 72.74(2.41) / \textbf{91.98(0.32)} / 82.76(2.40) & 78.00(1.95) / 87.90(1.83) / \textbf{97.80(0.15)} / 95.18(0.86) & 79.25(2.28) / 87.83(1.99) / \textbf{98.12(0.13)} / 95.92(0.68) \\
    & & iSUN & 58.54(1.43) / 71.72(2.57) / \textbf{91.69(0.46)} / 83.73(1.92) & 77.29(2.15) / 87.07(2.00) / \textbf{97.66(0.14)} / 95.39(0.55) & 80.27(2.59) / 88.13(1.99) / \textbf{98.18(0.10)} / 96.40(0.35) \\
    & & SVNH & 58.13(2.90) / 83.04(1.69) / \textbf{92.39(1.42)} / 90.28(1.57) & 79.82(2.49) / 93.46(1.05) / \textbf{97.96(0.49)} / 97.52(0.41) & 66.44(4.78) / 84.81(3.06) / \textbf{96.05(0.68)} / 95.13(0.54) \\
    & & Food-101 & 69.12(0.76) / 71.12(0.79) / 70.23(1.61) / \textbf{72.68(1.01)} & 89.25(0.40) / 90.76(0.35) / 91.15(0.66) / \textbf{92.53(0.38)} & 83.20(0.70) / 84.74(0.78) / 87.69(0.85) / \textbf{89.37(0.54)} \\
    & & STL-10 & 58.23(0.55) / 66.14(0.39) / \textbf{86.69(2.54)} / 71.03(1.17) & 77.68(0.61) / 83.02(1.12) / \textbf{94.73(1.36)} / 88.89(0.51) & 81.66(1.01) / 85.27(1.49) / \textbf{95.52(1.95)} / 91.70(0.43) \\
    & & MNIST & 61.79(5.62) / 87.27(4.78) / \textbf{94.15(3.68)} / 85.60(3.96) & 82.64(4.73) / 96.71(1.39) / \textbf{98.54(1.01)} / 96.03(1.49) & 85.99(3.87) / 97.18(1.17) / \textbf{98.89(0.65)} / 96.59(1.25) \\
    & & F-MNIST & 72.39(2.13) / 88.50(1.70) / 87.63(0.67) / \textbf{92.48(1.22)} & 90.33(1.02) / 96.61(0.68) / 96.61(0.30) / \textbf{98.27(0.36)} & 91.74(0.87) / 96.71(0.75) / 97.36(0.24) / \textbf{98.46(0.31)} \\
    & & NotMNIST & 56.15(1.61) / 81.37(2.36) / \textbf{96.48(0.38)} / 77.23(2.62) & 80.00(2.79) / 93.47(1.35) / \textbf{98.93(0.13)} / 93.05(1.47) & 73.99(4.65) / 89.37(2.60) / \textbf{98.84(0.15)} / 90.53(2.00) \\
    & & Gaussian & 47.54(0.07) / 86.50(21.83) / \textbf{97.50(0.00)} / \textbf{97.50(0.00)} & 59.59(30.02) / 80.04(41.24) / \textbf{100.00(0.00)} / 99.82(0.15) & 73.02(23.16) / 85.54(30.09) / \textbf{100.00(0.00)} / 99.88(0.09) \\
    & & Uniform & 47.51(0.02) / 75.44(20.61) / \textbf{97.50(0.00)} / \textbf{97.50(0.00)} & 60.88(14.02) / 92.66(8.46) / \textbf{100.00(0.00)} / 99.95(0.10) & 74.03(12.70) / 95.08(6.01) / \textbf{100.00(0.00)} / 99.97(0.07) \\
    \bottomrule
    \end{tabular}}
    \end{sc}
    \end{small}
    \end{center}
    % \vskip -0.2in
    \end{table}
    
\clearpage

\section {OOD Detection Using an Ensemble of Networks}

    The leave-out ensemble proposed in [38] uses multiple networks and is reported to achieve high OOD detection performance in the one-vs-one evaluation. 
    To make a fair comparison, we consider an extension of our method to an ensemble model. The underlying thought is that the use of an ensemble of multiple models will yield better results, as seen in many inference tasks. To be specific, in the training step, we train multiple networks on the target classification task; in our experiments, we trained models of the same architecture initialized with different random weights. At test time, given an input sample, we make the networks output the cosine similarities and calculate their averages over different networks. Table \ref{table:OOD_detection_result_ens} shows the results. For the leave-out ensemble, it shows the performances reported in [38] and those obtained in our own experiments (indicated by *); we used a public code\footnote{https://github.com/YU1ut/Ensemble-of-Leave-out-Classifiers} suggested by the author of [38].
    It is seen that our method shows better or at least comparable performance as compared with the leave-out ensemble, even in the one-vs-one evaluation. Note that iSUN is chosen as the validation OOD dataset for the hyperparameter determination of the leave-out ensemble, following [38].

    \begin{table}[h]
    \caption{
    Performance of OOD detection by  ensemble models (five networks) in the one-vs-one evaluation.     }
    \label{table:OOD_detection_result_ens}
    \begin{center}
    \begin{small}
    \begin{sc}
    \resizebox{0.60\textwidth}{!}{
    \begin{tabular}{cccccc}
    \toprule
    \multirow{2}{*}{Network} & \multirow{2}{*}{ID} & \multirow{2}{*}{OOD} & Accuracy & \multirow{2}{*}{AUROC} & \multirow{2}{*}{AUPR-In} \\
    & & & at TPR$=95\%$ & & \\\midrule
    & & & \multicolumn{3}{c}{Leave-out[38] / Leave-out* / Ours} \\ 
    \midrule
    \multirow{28}{*}{\rotatebox[origin=c]{0}{Dense-BC}} & \multirow{14}{*}{\rotatebox[origin=c]{0}{CIFAR-10}} & 
        TinyIm (c) &    96.89 / 96.76 / 96.43   &   99.65 / 99.66 / 99.43   &   99.68 / 99.67 / 99.48 \\
    & & TinyIm (c)* &   ~~~-~~ / 94.83 / 96.21   &   ~~~-~~ / 98.98 / 99.33   &   ~~~-~~ / 99.05 / 99.39 \\
    & & TinyIm (r) &    96.04 / 96.21 / 96.14   &   99.34 / 99.45 / 99.36   &   99.37 / 99.48 / 99.40 \\
    & & LSUN (c)   &    95.79 / 95.65 / 96.51   &   99.25 / 99.27 / 99.51   &   99.29 / 99.32 / 99.53 \\
    & & LSUN (c)*  &    ~~~-~~ / 91.04 / 96.06   &   ~~~-~~ / 97.83 / 99.33   &   ~~~-~~ / 98.00 / 99.36 \\
    & & LSUN (r)   &    97.12 / 96.71 / 96.79   &   99.75 / 99.67 / 99.61   &   99.77 / 99.68 / 99.64 \\
    & & iSUN       &    ~~~-~~ / 96.47 / 96.69   &   ~~~-~~ / 99.60 / 99.61   &   ~~~-~~ / 99.62 / 99.67 \\
    & & SVHN       &    ~~~-~~ / 81.09 / 96.53   &   ~~~-~~ / 94.39 / 99.54   &   ~~~-~~ / 95.06 / 98.93 \\
    & & Food-101   &    ~~~-~~ / 75.64 / 85.06   &   ~~~-~~ / 92.85 / 95.89   &   ~~~-~~ / 94.13 / 93.25 \\
    & & MNIST      &    ~~~-~~ / 97.19 / 97.01   &   ~~~-~~ /  99.76 / 99.53   &   ~~~-~~ /  99.79 / 99.61 \\
    & & F-MNIST    &    ~~~-~~ / 96.59 / 96.60   &   ~~~-~~ /  99.58 / 99.45   &   ~~~-~~ /  99.62 / 99.51 \\
    & & NotMNIST   &    ~~~-~~ / 93.39 / 97.11   &   ~~~-~~ /  98.64 / 99.45   &   ~~~-~~ /  98.83 / 99.24 \\
    & & Gausian    &    96.20 / 97.50 / 97.50   &   98.55 / 99.99 / 100.00  &   98.94 / 99.99 / 100.00 \\
    & & Uniform    &    97.50 / 97.50 / 97.50   &   99.84 / 99.96 / 100.00  &   99.89 / 99.97 / 100.00 \\ 
    \cmidrule(l){2-6}
    & \multirow{15}{*}{\rotatebox[origin=c]{0}{CIFAR-100}}&
        TinyIm (c) &    93.36 / 95.11 / 94.44   &   98.43 / 99.00 / 98.78   &   98.58 / 99.05 / 98.86 \\
    & & TinyIm (c)* &   ~~~-~~ / 88.78 / 93.09   &   ~~~-~~ / 96.79 / 98.30   &   ~~~-~~ / 97.03 / 98.46 \\
    & & TinyIm (r) &    87.24 / 91.45 / 93.89   &   96.27 / 97.80 / 98.60   &   96.66 / 98.01 / 98.72 \\
    & & LSUN (c)   &    90.16 / 89.34 / 91.54   &   97.37 / 97.05 / 97.80   &   97.62 / 97.26 / 97.91 \\
    & & LSUN (c)*  &    ~~~-~~ / 75.23 / 88.55   &   ~~~-~~ / 90.75 / 96.74   &   ~~~-~~ / 91.27 / 96.99 \\
    & & LSUN (r)   &    89.39 / 93.25 / 93.16   &   97.03 / 98.37 / 98.38   &   97.37 / 98.52 / 98.55 \\
    & & iSUN       &    ~~~-~~ / 90.91 / 92.93   &   ~~~-~~ / 97.53 / 98.23   &   ~~~-~~ / 97.66 / 98.53 \\
    & & SVHN       &    ~~~-~~ / 50.84 / 92.82   &   ~~~-~~ / 75.79 / 98.22   &   ~~~-~~ / 81.16 / 96.39 \\
    & & Food-101   &    ~~~-~~ / 74.44 / 76.42   &   ~~~-~~ / 92.39 / 93.76   &   ~~~-~~ / 93.82 / 91.23 \\
    & & STL-10     &    ~~~-~~ / 86.56 / 78.48   &   ~~~-~~ /  96.34 / 92.04   &   ~~~-~~ /  96.78 / 93.93 \\
    & & MNIST      &    ~~~-~~ / 96.35 / 89.23   &   ~~~-~~ /  99.27 / 97.35   &   ~~~-~~ /  99.39 / 97.68 \\
    & & F-MNIST    &    ~~~-~~ / 96.73 / 96.88   &   ~~~-~~ /  99.66 / 99.64   &   ~~~-~~ /  99.67 / 99.67 \\
    & & NotMNIST   &    ~~~-~~ / 92.20 / 87.92   &   ~~~-~~ /  98.07 / 96.97   &   ~~~-~~ /  98.25 / 95.57 \\
    & & Gausian    &    57.64 / 81.18 / 97.50   &   92.00 / 95.57 / 100.00  &   94.77 / 97.36 / 100.00 \\
    & & Uniform    &    78.24 / 95.72 / 97.50   &   94.89 / 97.73 / 100.00  &   96.36 / 98.61 / 100.00 \\ 
    \midrule
    \multirow{28}{*}{\rotatebox[origin=c]{0}{WRN-28-10}}&\multirow{14}{*}{\rotatebox[origin=c]{0}{CIFAR-10}}& 
        TinyIm (c) &    97.09 / 96.35 / 94.83   &   99.75 / 99.54 / 98.93   &   99.77 / 99.57 / 99.05 \\
    & & TinyIm (c)* &   ~~~-~~ / 93.47 / 94.62   &   ~~~-~~ / 98.56 / 98.77   &   ~~~-~~ / 98.65 / 98.95 \\
    & & TinyIm (r) &    96.03 / 95.09 / 93.41   &   99.36 / 99.10 / 98.41   &   99.40 / 99.13 / 98.62 \\
    & & LSUN (c)   &    96.54 / 94.97 / 96.42   &   99.55 / 99.09 / 99.46   &   99.57 / 99.16 / 99.49 \\
    & & LSUN (c)*  &    ~~~-~~ / 88.93 / 95.91   &   ~~~-~~ / 97.14 / 99.30   &   ~~~-~~ / 97.33 / 99.35 \\
    & & LSUN (r)   &    97.06 / 95.88 / 95.74   &   99.70 / 99.39 / 99.14   &   99.72 / 99.37 / 99.27 \\
    & & iSUN       &    ~~~-~~ / 95.33 / 95.53   &   ~~~-~~ / 99.22 / 99.06   &   ~~~-~~ / 99.22 / 99.28 \\
    & & SVHN       &    ~~~-~~ / 73.78 / 96.99   &   ~~~-~~ / 91.80 / 99.73   &   ~~~-~~ / 93.15 / 99.36 \\
    & & Food-101   &    ~~~-~~ / 66.64 / 81.60   &   ~~~-~~ / 87.37 / 95.10   &   ~~~-~~ / 89.23 / 92.34 \\
    & & MNIST      &    ~~~-~~ / 95.52 / 96.12   &   ~~~-~~ /  99.09 / 98.98   &   ~~~-~~ /  99.22 / 99.22 \\
    & & F-MNIST    &    ~~~-~~ / 96.59 / 95.33   &   ~~~-~~ /  99.59 / 99.04   &   ~~~-~~ /  99.60 / 99.16 \\
    & & NotMNIST   &    ~~~-~~ / 92.33 / 96.99   &   ~~~-~~ /  97.91 / 99.38   &   ~~~-~~ /  98.05 / 99.19 \\
    & & Gausian    &    89.31 / 97.50 / 97.50   &   96.77 / 99.97 / 100.00  &   97.78 / 99.98 / 100.00 \\
    & & Uniform    &    97.50 / 97.50 / 97.50   &   99.58 / 99.98 / 100.00  &   99.71 / 99.98 / 100.00 \\ 
    \cmidrule(l){2-6}
    & \multirow{15}{*}{\rotatebox[origin=c]{0}{CIFAR-100}} & 
        TinyIm (c) &    92.92 / 92.88 / 90.58   &   98.22 / 98.33 / 97.62   &   98.39 / 98.46 / 97.89 \\
    & & TinyIm (c)* &   ~~~-~~ / 84.19 / 87.83   &   ~~~-~~ / 95.34 / 96.80   &   ~~~-~~ / 95.69 / 97.26 \\
    & & TinyIm (r) &    85.24 / 88.74 / 87.09   &   95.18 / 96.89 / 96.64   &   95.50 / 97.16 / 97.11 \\
    & & LSUN (c)   &    90.39 / 83.34 / 88.61   &   97.38 / 95.43 / 97.00   &   97.62 / 96.03 / 97.26 \\
    & & LSUN (c)*  &    ~~~-~~ / 68.83 / 85.23   &   ~~~-~~ / 87.66 / 95.87   &   ~~~-~~ / 88.55 / 96.33 \\
    & & LSUN (r)   &    89.24 / 92.17 / 84.58   &   96.77 / 97.98 / 95.97   &   97.03 / 98.13 / 96.65 \\
    & & iSUN       &    ~~~-~~ / 90.33 / 85.66   &   ~~~-~~ / 97.31 / 96.17   &   ~~~-~~ / 97.41 / 97.05 \\
    & & SVHN       &    ~~~-~~ / 51.65 / 92.66   &   ~~~-~~ / 76.30 / 98.20   &   ~~~-~~ / 80.87 / 96.41 \\
    & & Food-101   &    ~~~-~~ / 68.62 / 78.00   &   ~~~-~~ / 89.87 / 94.38   &   ~~~-~~ / 91.49 / 92.02 \\
    & & STL-10     &    ~~~-~~ / 80.86 / 72.86   &   ~~~-~~ /  94.33 / 90.16   &   ~~~-~~ /  95.00 / 92.72 \\
    & & MNIST      &    ~~~-~~ / 93.84 / 86.94   &   ~~~-~~ /  98.62 / 96.79   &   ~~~-~~ /  98.82 / 97.23 \\
    & & F-MNIST    &    ~~~-~~ / 97.05 / 95.37   &   ~~~-~~ /  99.77 / 99.16   &   ~~~-~~ /  99.79 / 99.24 \\
    & & NotMNIST   &    ~~~-~~ / 86.31 / 78.95   &   ~~~-~~ /  96.53 / 94.32   &   ~~~-~~ /  97.03 / 92.40 \\
    & & Gausian    &    47.55 / 97.50 / 97.50   &   83.44 / 99.65 / 99.98   &   89.43 / 99.79 / 99.99 \\
    & & Uniform    &    48.37 / 97.50 / 97.50   &   93.04 / 99.60 / 100.00  &   88.64 / 99.76 / 100.00 \\ 
    \bottomrule
    \end{tabular}}
    \end{sc}
    \end{small}
    \end{center}
    \end{table}

\clearpage
    
\section{Black Frame in the Cropped OOD Images} \label{sec:black-frame-explain}

\begin{figure}[h]
    \centerline{\includegraphics[width=9.2cm]{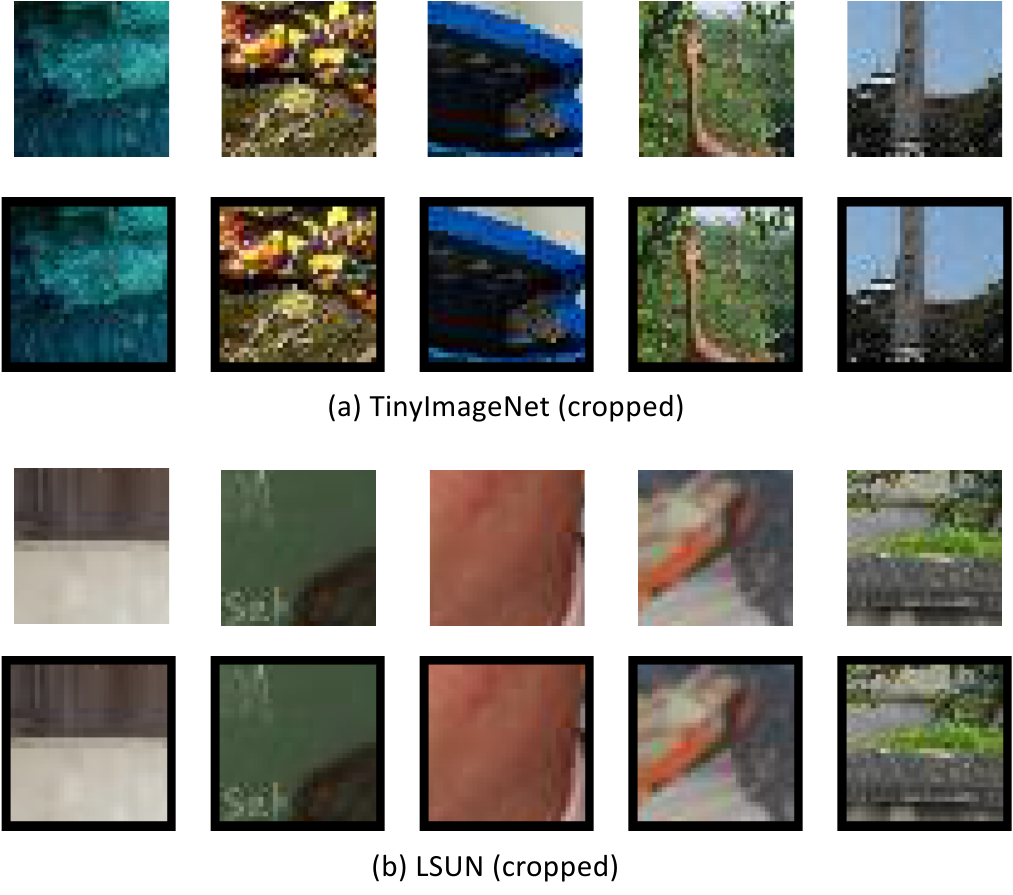}}
    \caption{The images with and without a black frame. (a) TinyImageNet (cropped). (b) LSUN (cropped).}
    \label{fig:black-frame}
    \end{figure}

    The OOD datasets, i.e., LSUN (cropped \& resized), TinyImageNet (cropped \& resized), and iSUN, provided in the authors' GitHub repo of ODIN [25] \footnote{https://github.com/facebookresearch/odin} are used in many studies [25, 38, 45]. As mentioned briefly in the main paper, we found that every image in the datasets of cropped images, i.e., Tiny ImageNet (cropped) and LSUN (cropped), unexpectedly has a black frame with two-pixel widths, as shown in Fig.~\ref{fig:black-frame}. Those images is of $36\times 36$ pixels (4 pixels larger than CIFAR images), implying that it is a mistake of the authors. In any case, adding a black frame is not invalid by itself, as any image could be an OOD sample. However, it will ease the problem without a doubt. 
        
    In our experiments, we used both of the corrected versions (indicated by *) and the original version.
    The results show that the proposed method achieves similar performance on both versions of the datasets. On the other hand, the other methods, ODIN [25], Mahalanobis detector [23] and Leave-out Ensemble [38], show more sensitive behaviors to the difference, as observable in Tables 5 - 8 in this article.  

\end{document}